\documentclass[runningheads]{llncs}
\usepackage{graphicx}
\usepackage{amsmath,amssymb} %
\usepackage{color}
\usepackage{booktabs,arydshln}
\usepackage[font=small,labelfont=bf]{caption}
\makeatletter
\def\adl@drawiv#1#2#3{%
        \hskip.5\tabcolsep
        \xleaders#3{#2.5\@tempdimb #1{1}#2.5\@tempdimb}%
                #2\z@ plus1fil minus1fil\relax
        \hskip.5\tabcolsep}
\newcommand{\cdashlinelr}[1]{%
  \noalign{\vskip\aboverulesep
           \global\let\@dashdrawstore\adl@draw
           \global\let\adl@draw\adl@drawiv}
  \cdashline{#1}
  \noalign{\global\let\adl@draw\@dashdrawstore
           \vskip\belowrulesep}}
\makeatother

\usepackage{multirow}
\usepackage{tabularx}
\usepackage{xspace}
\usepackage{mathtools}
\belowdisplayskip \abovedisplayskip

\newlength{\sectionReduceTop}
\newlength{\sectionReduceBot}
\newlength{\subsectionReduceTop}
\newlength{\subsectionReduceBot}
\newlength{\abstractReduceTop}
\newlength{\abstractReduceBot}
\newlength{\captionReduceTop}
\newlength{\captionReduceBot}
\newlength{\subsubsectionReduceTop}
\newlength{\subsubsectionReduceBot}

\newlength{\eqnReduceTop}
\newlength{\eqnReduceBot}

\newlength{\horSkip}
\newlength{\verSkip}

\newlength{\figureHeight}
\makeatletter
\DeclareRobustCommand\onedot{\futurelet\@let@token\@onedot}
\def\@onedot{\ifx\@let@token.\else.\null\fi\xspace}
\def\eg{e.g\onedot} 
\def\ie{i.e\onedot} 
 
\def\etc{etc\onedot} 
 
\def\etal{\textit{et~al\onedot}} 
\def\Fig{Fig\onedot} 
\def\Table{Tab\onedot} 

\def\Sec{Sec\onedot}

\makeatother

\newcommand{\nmn}{CorefNMN\xspace}

\newcommand{\reftab}[1]{Tab.~\ref{#1}}

\newcommand{\myparagraph}[1]{\vspace{0pt}\noindent{\bf #1}}

\newcommand{\addresults}[1]{}

\begin{document}
\title{Visual Coreference Resolution in Visual Dialog using Neural Module Networks} 
\titlerunning{Visual Coreference Resolution in Visual Dialog}

\author{Satwik Kottur\inst{1,2}\thanks{Work partially done as an intern at Facebook AI Research} \and
Jos\'{e} M. F. Moura\inst{2} \and
Devi Parikh \inst{1,3} \and
Dhruv Batra \inst{1,3} \and
Marcus Rohrbach  \inst{1}}
\authorrunning{S. Kottur, J. M. F. Moura, D. Parikh, D. Batra, M. Rohrbach}
\institute{
Facebook AI Research, Menlo Park, USA \and
Carnegie Mellon University, Pittsburgh, USA \and
Georgia Institute of Technology, Atlanta, USA\\
}
\maketitle
\begin{abstract}
Visual dialog entails answering a series of questions grounded in an image, 
using dialog history as context.
In addition to the challenges found in visual question answering (VQA),
which can be seen as one-round dialog, visual dialog encompasses several more.
We focus on one such problem called \textit{visual coreference resolution} that
involves determining which words, typically noun phrases and pronouns, 
\textit{co-refer} to the same entity/object instance in an image.
This is crucial, especially for pronouns (\eg, `\textit{it}'), as the dialog agent
must first link it to a previous coreference (\eg, \textit{`boat'}), and only then
can rely on the visual grounding of the coreference \textit{`boat'} to reason about the
pronoun `\textit{it}'.
Prior work (in visual dialog) models visual coreference resolution either
(a) implicitly via a memory network over history, or 
(b) at a coarse level for the entire question;
and not explicitly at a phrase level of granularity.
In this work, we propose a neural module network architecture for visual dialog
by introducing two novel modules---\texttt{Refer} and \texttt{Exclude}---that
perform explicit, grounded, coreference resolution at a finer word level.
We demonstrate the effectiveness of our model on MNIST Dialog, a visually 
simple yet coreference-wise complex dataset,
by achieving near perfect accuracy, and on VisDial, a large and
challenging visual dialog dataset on real images, where our model 
outperforms other approaches, and is more interpretable, grounded, and 
consistent qualitatively.
\end{abstract}

\section{Introduction}
\begin{figure}[t]
	\centering
    \includegraphics[width=0.95\textwidth]{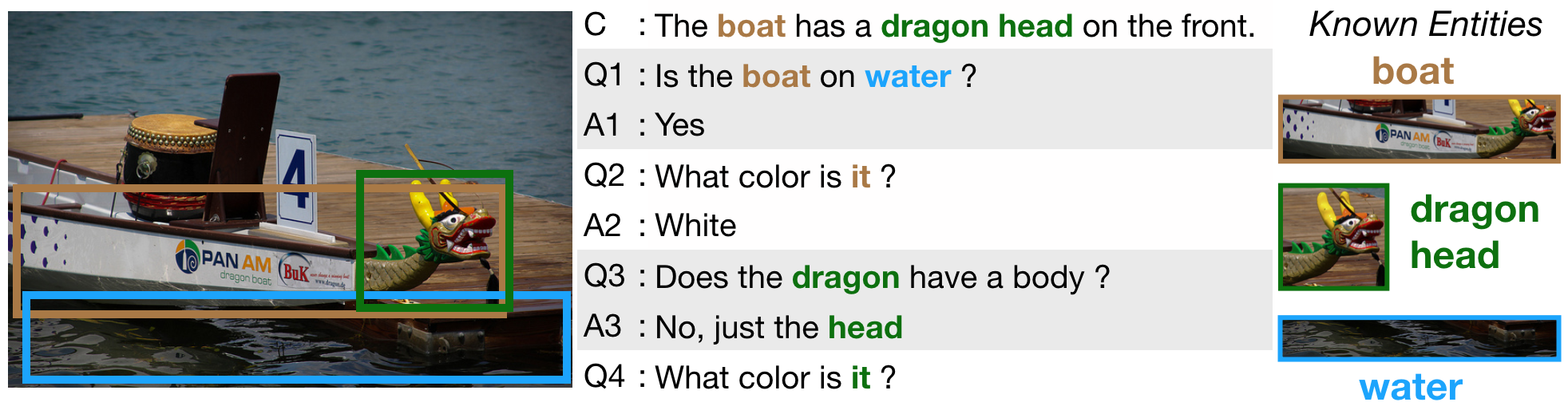}
    \caption{
    Our model begins by grounding entities in the caption (C), \textit{boat} 
    (brown) and \textit{dragon head} (green), and stores them in a pool for 
    future coreference resolution in the dialog (right).
    When asked \textit{`Q1: Is the boat on water?'}, it identifies that
    the \textit{boat} (known entity) and \textit{water} (unknown entity) 
    are crucial to answer the question.
    It then grounds the novel entity \textit{water} in the image (blue), but 
    resolves \textit{boat} by referring back to the pool and reusing the 
    available grounding from the caption, before proceeding with further 
    reasoning.
    Thus, our model explicitly resolves coreferences in visual dialog.} 
    \label{fig:teaser}
\end{figure}

The task of Visual Dialog \cite{visdial,guesswhat} involves building agents 
that `see' (\ie understand an image) and `talk' (\ie communicate this 
understanding in a dialog).
Specifically, it requires an agent to answer a sequence of questions about an 
image, requiring it to reason about both the image and the past dialog history.
For instance, in \Fig\ref{fig:teaser}, to answer \textit{`What color is it?'},
the agent needs to reason about the history to know what \textit{`it'} refers
to and the image to find out the color.
This generalization of visual question answering (VQA) \cite{antol15iccv} to
dialog takes a step closer to real-world applications (aiding visually impaired 
users, intelligent home assistants, natural language interfaces for robots) but
simultaneously introduces new modeling challenges at the intersection of vision
and language. 
The particular challenge we focus on in this paper is that of \emph{visual 
coreference resolution} in visual dialog. 
Specifically, we introduce a new model that performs explicit visual 
coreference resolution and interpretable entity tracking in visual dialog. 

It has long been understood 
\cite{grice1975logic,Winograd:1972:UNL:540414,mitchell-vandeemter-reiter:2013:NAACL-HLT,yu16eccv} 
that humans use \emph{coreferences}, different phrases and short-hands such as pronouns, to refer to the same entity or referent in a single text. %
In the context of visually grounded dialog, we are interested in referents which are in the image, e.g. an object or person. All phrases in the dialog which refer to the same entity or referent in the image are called visual coreferences.
Such coreferences can be noun phrases such as \emph{`a dragon head'}, 
\emph{`the head'}, or pronouns such as \emph{`it'} (\Fig\ref{fig:teaser}).
Especially when trying to answer a question that contains an anaphora, for
instance the pronoun \emph{`it'}, which refers to its full form (the antecedent) \emph{`a dragon head'}, it is necessary to \textit{resolve} the coreference 
on the language side and ground it to the underlying visual referent.
More specifically, to answer the question \textit{`What color is it?'} in
\Fig\ref{fig:teaser}, the model must correctly identify which object 
\textit{`it'} refers to, in the given context.
Notice that a word or phrase can refer to different entities in 
different contexts, as is the case with \textit{`it'} in this example.
Our approach to explicitly resolve visual coreferences is inspired from the 
functionality of variables or memory in a computer program.
In the same spirit as how one can refer back to the contents of variables at a 
later time in a program without explicitly re-computing them, we propose a 
model which can refer back to entities from previous rounds of dialog and 
reuse the associated information; and in this way resolve coreferences.

Prior work on VQA \cite{malinowski15iccv,fukui16emnlp,anderson18cvpr} has (understandably) largely ignored the problem of visual coreference resolution since individual questions asked in isolation rarely contain coreferences.
In fact, recent empirical studies 
\cite{AgrawalBP16,JohnsonHMFZG16,ZhangGSBP15,goyal17cvpr} suggest that today's 
vision and language models seem to be exploiting dataset-level statistics and 
perform poorly at grounding entities into the correct pixels. 
In contrast, our work aims to explicitly reason over past 
dialog interactions by referring back to previous references.
This allows for increased interpretability of the model.
As the dialog progresses (\Fig\ref{fig:teaser}), we can inspect the pool of
entities known to the model, and also visualize which entity a particular
phrase in the question has been resolved to.
Moreover, our explicit entity tracking model has benefits even in cases that 
may not strictly speaking require coreference resolution.
For instance, by explicitly referring \textit{`dragon'} in Q3 
(\Fig\ref{fig:teaser}) back to a known entity, the model 
is consistent with itself and (correctly) grounds the phrase in the image. 
We believe such consistency in model outputs is a strongly desirable property 
as we move towards human-machine interaction in dialog systems.%

Our main technical contribution is a neural module network architecture for 
visual dialog.
Specifically, we propose two novel modules---\texttt{Refer} and 
\texttt{Exclude}---that
perform explicit, grounded, coreference resolution in visual dialog.
In addition, we propose a novel way to handle captions using
neural module networks at a word-level granularity finer than a traditional 
sentence-level encoding.
We show quantitative benefits of these modules on a reasoning-wise 
complicated but visually simple MNIST dialog dataset \cite{paul2017visual}, 
where achieve near perfect accuracy.
On the visually challenging VisDial dataset \cite{visdial}, our model not only
outperforms other approaches but also is more 
interpretable by construction and enables word-level coreference resolution.
Furthermore, we qualitatively show that our model is
(a) more interpretable (a user can inspect which entities were detected and
tracked as the dialog progresses, and which ones were referred to for answering
a specific question),
(b) more grounded (where the model looked to answer a question in the dialog),
(c) more consistent (same entities are considered across rounds of dialog).

\section{Related Work}
We discuss:
(a) existing approaches to visual dialog, 
(b) related tasks such as visual grounding and coreference resolution, and
(c) neural module networks.

\myparagraph{Visual Dialog.}
Though the origins of visual dialog can be traced back to \cite{winograd1971procedures,geman_pnas14},
it was largely formalized by \cite{visdial,guesswhat} who collected 
human annotated datasets for the same.
Specifically, \cite{visdial} paired annotators to collect free-form 
natural-language questions and answers, where the questioner was instructed to 
ask questions to help them imagine the hidden scene (image) better.
On the other hand, dialogs from \cite{guesswhat} are more goal driven
and contain yes/no questions directed towards identifying a secret object in 
the image.
The respective follow up works used reinforcement learning techniques to solve
this problem \cite{visdial_rl,DBLP:journals/corr/StrubVMPCP17}.
Other approaches to visual dialog include transferring knowledge from a 
discriminatively trained model to a generative dialog model \cite{lu_nips16}, 
using attention networks to solve visual coreferences \cite{paul2017visual},
and more recently, a probabilistic treatment of dialogs using conditional
variational autoencoders \cite{1802.03803}.
Amongst these, \cite{paul2017visual} is the closest to this work, while
\cite{lu_nips16,1802.03803} are complementary.
To solve visual coreferences, \cite{paul2017visual} relies on global visual 
attentions used to answer previous questions.
They store these attention maps in a memory against keys based on textual 
representations of the entire question and answer, along with the history.
In contrast, operating at a finer word-level granularity within each 
question, our model can resolve different phrases of a question, and 
ground them to different parts of the image, a core component in correctly 
understanding and grounding coreferences.
E.g.,  \textit{`A man and woman in a car. Q: Is he or she 
driving?'}, which requires resolving \textit{`he'} and \textit{`she'}
individually to answer the question.

\myparagraph{Grounding language in images and video.}
Most works in this area focus on the specific task of localizing a textual 
referential expression in the image 
\cite{hu16cvpr,kong14cvpr,mao16cvpr,plummer15iccv,rohrbach16eccv,wang2016cvpr,yu16eccv}
or video
\cite{regneri13tacl,lin14cvpr,yu13acl,hendricks17iccv}.
Similar to these works, one component of our model aims to localize words and 
phrases in the image.
However, the key difference is that if the phrase being grounded is an anaphora
(\eg, \textit{`it'}, \textit{`he'}, \textit{`she'}, \etc),
our model first resolves it explicitly to a known entity, and then grounds it
by borrowing the referent's visual grounding.

\myparagraph{Coreference resolution.}
The linguistic community defines coreference resolution as the task of
clustering phrases, such as noun phrases and pronouns, which refer to the same entity in the world (see, for example, \cite{bergsma06acl}). 
The task of visual coreference resolution  links the coreferences to an entity in 
the visual data.
For example, \cite{ramanathan14eccv} links character mentions in TV show 
descriptions with their occurrence in the video,
while \cite{kong14cvpr} links text phrases to objects in a 3D 
scene.
Different from these works, we predict a program for a given natural 
language question about an image, which then tries to resolve any existing 
coreferences, to then answer the question. 
An orthogonal direction is to generate language while jointly grounding and 
resolving coreferences -- \eg, \cite{rohrbach17cvpr} explore this for movie 
descriptions.
While out of scope for this work, it is an interesting direction for future 
work in visual dialog, especially when generating questions.

\myparagraph{Neural Module Networks} \cite{andreas16cvpr}
are an elegant class of models where an instance-specific architecture is 
composed from neural `modules' (or building blocks) that are shared across instances.
The high-level idea is inspired by `options' or sub-tasks in hierarchical RL. 
They have been shown to be successful for visual question answering in real 
images and linguistic databases \cite{andreas_naacl16} and for more complex 
reasoning tasks in synthetic datasets \cite{johnson17iccv,hu2017learning}.
For this, \cite{johnson17iccv,hu2017learning} learn program prediction and 
module parameters jointly, end-to-end.
Within this context, our work generalizes the formulation in 
\cite{hu2017learning} from VQA to visual dialog by introducing a novel module to perform explicit visual coreference resolution.

\section{Approach}
\label{sec:approach}

\begin{figure}[t]
	\centering
    \includegraphics[width=0.9\textwidth]{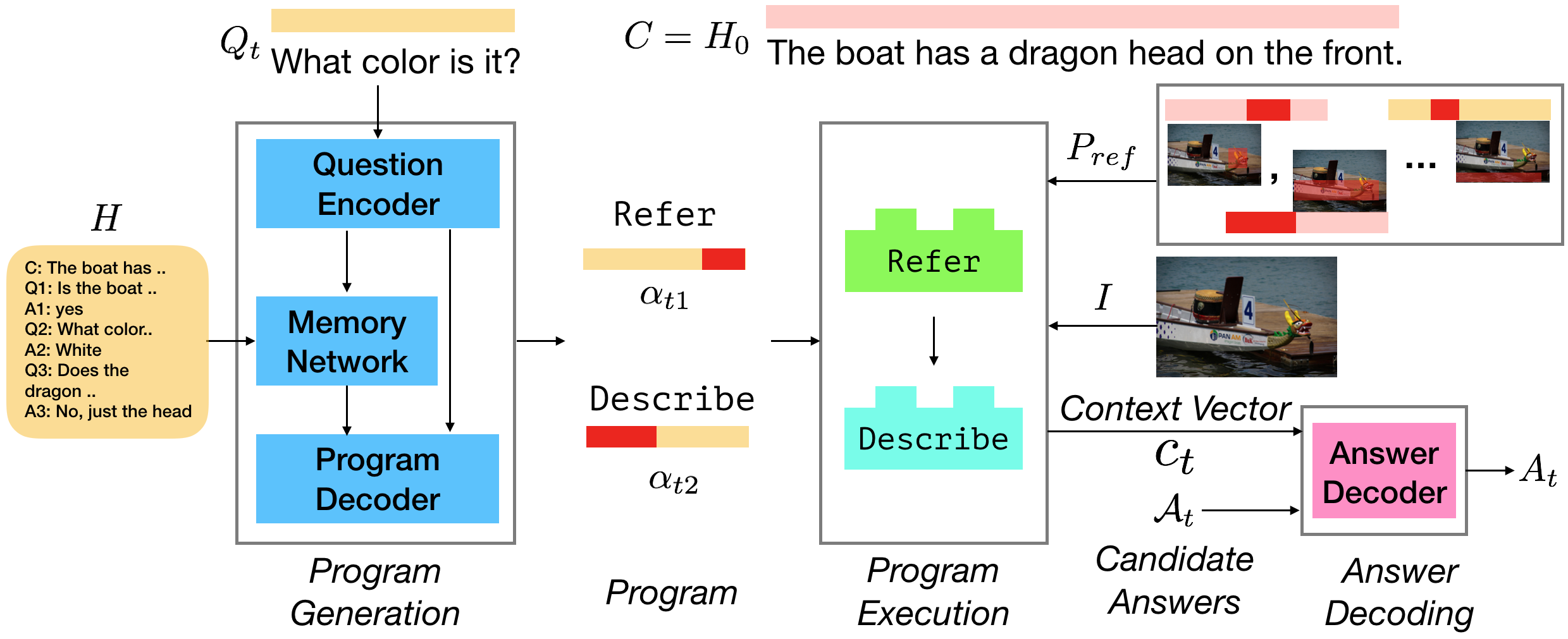}
    \caption{
    Overview of our model architecture.
    The question $Q_t$ (orange bar) is encoded along with the history $H$ 
    through a memory augmented question encoder, using which a program 	
    (\texttt{Refer Describe}) is decoded.
    For each module in the program, an attention $\alpha_{ti}$ over $Q_t$ is 
    also predicted, used to compute the text feature $x_{txt}$.
	For $Q_t$, attention is over \textit{`it'} for \texttt{Refer} and 
    \textit{`What color'} for \texttt{Describe}, respectively
  	(orange bars with red attention).
    \texttt{Refer} module uses the coreference pool $P_{ref}$, a dictionary of
    all previously seen entities with their visual groundings,
    resolves \textit{`it'}, and borrows the referent's visual grounding 	
    (\textit{boat} in this case).
    Finally, \texttt{Describe} extracts the `color' to produce $c_t$ used by a 
    final decoder to pick the answer $A_t$ from the candidate pool 
    $\mathcal{A}_t$.
    }
    \label{fig:model_fig}
\end{figure}

Recall that visual dialog \cite{visdial} involves answering a
question $Q_t$ at the current round $t$, given an image $I$, and the
dialog history (including the image caption) $H = (\underbrace{C}_{H_0},
    \underbrace{(Q_1, A_1)}_{H_1}, \cdots,
    \underbrace{(Q_{t-1}, A_{t-1})}_{H_{t-1}})$, by ranking a list of 
$100$ candidate answers 
$\mathcal{A}_t = \{A_t^{(1)},\cdots,A_t^{(100)}\}$.
As a key component for building better visual dialog agents, our model
explicitly resolves visual coreferences in the current question, if
any.

Towards this end, our model first identifies relevant words or phrases in the 
current question that refer to entities in the image (typically objects and 
attributes).
The model also predicts whether each of these has been mentioned in the dialog
so far.
Next, if these are novel entities (unseen in the dialog history), they are 
localized in the image before proceeding, and for seen entities, the model 
predicts the (first) relevant coreference in the conversation history, and retrieves its corresponding visual grounding.
Therefore, as rounds of dialog progress, the model collects unique entities and
their corresponding visual groundings, and uses this \textit{reference pool} to
resolve any coreferences in subsequent questions.

Our model has three broad components:
(a) \textit{Program Generation} (\Sec\ref{sec:prog_predict}), where a reasoning
pathway, as dictated by a \textit{program},
is predicted for the current question $Q_t$,
(b) \textit{Program Execution} (\Sec\ref{sec:prog_execute}), where the
predicted program is executed by dynamically connecting neural modules
\cite{andreas_naacl16,andreas16cvpr,hu2017learning} to produce a
\textit{context} vector
summarizing the semantic information required to answer $Q_t$ from the context
($I, H$), and lastly,
(c) \textit{Answer Decoding} (\Sec\ref{sec:prog_execute}), where the
context vector $c_t$ is used to obtain the final answer $\hat{A}_t$.
We begin with a general characterization of neural modules used for VQA in \Sec\ref{sec:appraoch:modules:vqa} and then discuss our
novel modules for coreference resolution (\Sec\ref{sec:appraoch:modules:coref}) with details of the reference pool.
After describing the inner working of the modules, we explain each 
of the above three components of our model.
\subsection{Neural Modules for Visual Question Answering}
\label{sec:appraoch:modules:vqa}

The main technical foundation of our model is the neural module network (NMN) \cite{andreas16cvpr}.
In this section, we briefly recap NMNs and more specifically, the attentional
modules from \cite{hu2017learning}.
In the next section, we discuss novel modules we propose to handle additional 
challenges in visual dialog.

For a module $m$, let $x_{vis}$ and $x_{txt}$ be the input image and text 
embeddings, respectively.
In particular, the image embeddings $x_{vis}$ are spatial activation
maps of the image $I$ from a convolutional neural network.
The text embedding $x_{txt}$ is computed as a weighted sum of embeddings
of words in the question $Q_t$ using the soft attention weights $\alpha$
predicted by a program generator for module $m$ (more details in 
\Sec\ref{sec:prog_predict}).
Further, let $\{a_i\}$ be the set of $n_m$ single-channel spatial maps 
corresponding to the spatial image embeddings, where $n_m$ is the number of
attention inputs to $m$.
Denoting the module parameters with $\theta_m$, a neural module $m$ is 
essentially a parametric function 
$y = f_m(x_{vis}, x_{txt}, \{a_i\}_{i=1}^{n_m}; \theta_m)$.
The output from the module $y$ can either be a spatial image attention map 
(denoted by $a$) or a context vector (denoted by $c$), depending on the module.
The output spatial attention map $a$ feeds into next level modules while a context 
vector $c$ is used to obtain the final answer $A_t$.
The upper part of \Table\ref{tab:module_table} lists modules we adopt from 
prior work, with their functional forms.
We shortly summarize their behavior.
A \texttt{Find} module localizes objects or attributes by producing an 
attention over the image.
The \texttt{Relocate} module takes in an input image attention and performs
necessary spatial relocations to handle relationships like \textit{`next to'},
\textit{`in front of'}, \textit{`beside'}, \etc.
Intersection or union of attention maps can be obtained using \texttt{And} and \texttt{Or}, respectively.
Finally, \texttt{Describe}, \texttt{Exist}, and \texttt{Count} input an 
attention map to produce the context vector by describing an attribute, 
checking for existence, or counting, respectively, in the given input attention
map.
As noted in \cite{hu2017learning}, these modules are designed and named for
a potential `atomic' functionality.
However, we do not enforce this explicitly and let the modules discover their
expected behavior by training in an end-to-end manner.

\begin{table}[t]
	\setlength{\tabcolsep}{10pt}
	\centering
    {\resizebox{\columnwidth}{!}{%
    \begin{tabularx}{\textwidth}{lccl}
    \toprule
    \textbf{Name} & \textbf{Inputs} & \textbf{Output} & \textbf{Function}\\
    \midrule
    \multicolumn{4}{c}{\textbf{Neural Modules for VQA \cite{hu2017learning}}}\\
\texttt{Find} & $x_{vis}, x_{txt}$ & attention &
		$y = \text{conv}_2(\text{conv}_1(x_{vis} \odot Wx_{txt}))$\\
    
    \multirow{2}{*}{\texttt{Relocate}} &
        \multirow{2}{*}{$a, x_{vis}, x_{txt}$} & 
        \multirow{2}{*}{attention} & 
		$\tilde{y} = W_1 \text{sum}(a \odot x_{vis})$\\
   
   		& & & 
    	$y = \text{conv}_2(\text{conv}_1(x_{vis})
        		\odot \tilde{y} \odot W_2 x_{txt})$\\
    
    \texttt{And} & $a_1, a_2$ & attention &
		$y = \min\{a_1, a_2\}$\\
        
	\texttt{Or} & $a_1, a_2$ & attention &
		$y = \max\{a_1, a_2\}$\\

	\texttt{Exist} & $a, x_{vis}, x_{txt}$ & context &
		$y = W^T \text{vec}(a)$\\

	\texttt{Describe} & $a, x_{vis}, x_{txt}$ & context &
		$y = W_1^T(W_2 \text{sum}(a \odot x_{vis}) \odot W_3 x_{txt})$\\
        
    \texttt{Count} & $a, x_{vis}, x_{txt}$ & context &
		$y = W_1^T([\text{vec}(a), \max\{a\}, \min\{a\}])$\\
    	
\midrule
\multicolumn{4}{c}{\textbf{Neural Modules for Coreference resolution (Ours)}}\\
	\texttt{Not} & $a$ & attention &
		$y = \text{norm}_{L_1}(1 - a)$\\
        
	\texttt{Refer} & $ x_{txt}, P_{ref}$ & attention &
		(see text for details, \eqref{eq:refer-score-3})\\

	\texttt{Exclude} & $a, x_{vis}, x_{txt}$ & attention &
		$y =$ \texttt{And}[\texttt{Find}[$x_{vis}, x_{txt}$], 
        					\texttt{Not}[$a$]]\\
    \bottomrule
    \end{tabularx}%
    }}
    \vspace*{3pt}
    \caption{
    Neural modules used in our work for visual dialog, along with their 
    inputs, outputs, and function formulations.
    The upper portion contains modules from prior work used for visual question
    answering, while the bottom portion lists our novel modules designed to 
    handle additional challenges in visual dialog.}
    \label{tab:module_table}
\end{table}
\subsection{Neural Modules for Coreference Resolution}
\label{sec:appraoch:modules:coref}
We now introduce novel components and modules to handle visual dialog.

\myparagraph{Reference Pool ($P_{ref}$).}
The role of the reference pool is to keep track of entities seen so far
in the dialog.
Thus, we design $P_{ref}$ to be a dictionary of key-value pairs 
$(x_{txt}, a)$ 
for all the \texttt{Find} modules instantiated while answering
previous questions $(Q_i)_{i=1}^{t-1}$.
Recall that \texttt{Find} localizes objects/attributes specified by 
$x_{txt}$, and thus by storing each output attention map $y$, we now have 
access 
to all the entities mentioned so far in the dialog with their corresponding
visual groundings.
Interestingly, even though $x_{txt}$ and $y$ are intermediate 
outputs from our model, both are easily interpretable, making our reference
pool a \emph{semantic dictionary}.
To the best of our knowledge, our model is the first to attempt explicit,
interpretable coreference resolution in visual dialog.
While \cite{paul2017visual} maintains a dictionary similar to $P_{ref}$,
they do not consider word/entity level coreferences nor do their keys lend 
similar interpretability as ours.
With $P_{ref} = \{(x_p^{(i)}, a_p^{(i)})\}_i$ as input to \texttt{Refer},
we can now resolve references in $Q_t$.

\myparagraph{\texttt{Refer} Module.}
This novel module is responsible for 
resolving references in the question $Q_t$ and ground them in the conversation 
history $H$.
To enable grounding in dialog history, we generalize the above formulation
 to give the module access to a pool of references $P_{ref}$ of previously 
 identified entities.
Specifically, \texttt{Refer} only takes the text embedding $x_{txt}$ and the 
reference pool $P_{ref}$ as inputs, and resolves the entity represented by 
$x_{txt}$ in the form of a soft attention $\alpha$ over $Q_t$.
in this section after introducing $P_{ref}$.
For the example shown in \Fig\ref{fig:model_fig},
$\alpha$ for \texttt{Refer} attends to \textit{`it'}, indicating the phrase it
is trying to resolve.

At a high level, \texttt{Refer} treats $x_{txt}$ as a `query' and retrieves
the most likely match from $P_{ref}$ as measured by some similarity with
respect to keys $\{x_p^{(i)}\}_i$ in $P_{ref}$.
The associated image attention map of the best match is used as the visual 
grounding for the phrase that needed resolution (\ie \textit{`it'}).
More concretely, we first learn a \textit{scoring network} which when given a 
query $x_{txt}$ and a possible candidate $x_p^{(i)}$, returns a scalar value 
$s_i$ indicating how likely these text features refer to the same entity
\eqref{eq:refer-score-1}.
To enable \texttt{Refer} to consider the sequential nature of dialog when 
assessing a potential candidate, we additionally provide $\Delta_i t$, a measure of the `distance' of a candidate $x_p^{(i)}$ from $x_{txt}$ 
in the dialog history, as input to the scoring network.
$\Delta_i t$ is formulated as the absolute difference between the round of
$x_{txt}$ (current round $t$) and the round when $x_p^{(i)}$ was first
mentioned.
Collecting these scores from all the candidates, we apply a softmax function
to compute contributions $\tilde{s}_i$ from each entity in the pool 
\eqref{eq:refer-score-2}.
Finally, we weigh the corresponding attention maps via these contributions
to obtain the visual grounding $a_{out}$ for $x_{txt}$
\eqref{eq:refer-score-3}.
\begin{multicols}{2}
\vspace*{-30pt}
\begin{align}
s_i &= \text{MLP}([x_{txt}, x_p^{(i)}, \Delta_i t]) \label{eq:refer-score-1}\\
\tilde{s}_i &= \text{Softmax}(s_i) \label{eq:refer-score-2}
\end{align}
\columnbreak
\begin{align}
a_{out} &= \sum_{i=1}^{|P_{ref}|} \tilde{s}_i a_p^{(i)} \label{eq:refer-score-3}
\end{align}
\end{multicols}
\vspace*{-20pt}

\myparagraph{\texttt{Not} Module.}
Designed to focus on regions of the image \textbf{not} attended by the input 
attention map $a$, it outputs $y = \text{norm}_{L_1}(1-a)$, where 
norm$_{L_1}(.)$ normalizes the entries to sum to one.
This module is used in \texttt{Exclude}, described next.

\myparagraph{\texttt{Exclude} Module.}
To handle questions like \textit{`What other red things are present?'}, which 
seek other objects/attributes in the image than those specified by an input
attention map $a$, we introduce yet another novel module -- \texttt{Exclude}.
It is constructed using \texttt{Find}, \texttt{Not}, and \texttt{And} modules
as $y =$ \texttt{And}[\texttt{Find}[$x_{txt}, x_{vis}$], \texttt{Not}[$a$]],
where $x_{txt}$ is the text feature input to the \texttt{Exclude} module, for 
example, \textit{`red things'}.
More explicitly, \texttt{Find} first localizes all objects instances/attributes 
in the image.
Next, we focus on regions of the image other than those specified by $a$ using 
\texttt{Not}[$a$].
Finally, the above two outputs are combined via \texttt{And} to obtain the 
output $y$ of the \texttt{Exclude} module.

\subsection{Program Generation}
\label{sec:prog_predict}
A \textit{program} specifies the network layout for the neural modules 
for a given question $Q_t$.
Following \cite{hu2017learning}, it is serialized through the reverse 
polish notation (RPN) \cite{10.2307/2001990}.
This serialization helps us convert a hard, structured prediction problem into
a more tractable sequence prediction problem.
In other words, we need a program predictor to output a series of module
tokens in order, such that a valid layout can be retrieved from it.
There are two primary design considerations for our predictor.
First, in addition to the program, our predictor must also output a
soft attention $\alpha_{ti}$, over the question $Q_t$, for every module 
$m_i$ in the program.
This attention is responsible for the \emph{correct} module instantiation
in the current context.
For example, to answer the question \textit{`What color is the cat sitting next
to the dog?'}, a \texttt{Find} module instance attending to \textit{`cat'}
qualitatively serves a different purpose than one attending to \textit{`dog'}.
This is implemented by using the attention over $Q_t$ to compute the text 
embedding $x_{txt}$ that is directly fed as an input to the module during 
execution.
Second, to decide whether an entity in $Q_t$ has been seen before in the
conversation, it must be able to `peek' into the history $H$.
Note that this is unique to our current problem and does not exist in \cite{hu2017learning}.
To this effect, we propose a novel augmentation of attentional recurrent neural 
networks~\cite{bahdanau_iclr15} with memory \cite{DBLP:journals/corr/WestonCB14}
to address both the requirements (\Fig\ref{fig:model_fig}).

The program generation proceeds as follows.
First, each of the words in $Q_t$ are embedded to give $\{w_{ti}\}_{i=1}^T$,
where $T$ denotes the number of tokens in $Q_t$.
We then use a \textit{question encoder}, a multi-layer LSTM, to process
$w_{ti}$'s, resulting in a sequence of hidden states $\{\hat{w}_{ti}\}_{i=1}^T$
\eqref{eq:ques-enc-1}.
Notice that the last hidden state $h_T$ is the question encoding, which we
denote with $q_t$.
Next, each piece of history $(H_i)_{i=0}^{t-1}$ is processed in a similar
way by a \textit{history encoder}, which is a multi-layer LSTM akin to the
question encoder.
This produces encodings $(h_i)_{i=0}^{t-1}$ \eqref{eq:hist-mem-1} that serve 
as memory units to help the program predictor `peek' into the conversation
history.
Using the question encoding $q_t$, we attend over the history encodings 
$(h_i)_{i=0}^{t-1}$, and obtain the history vector $\hat{h}_t$ 
\eqref{eq:hist-mem-2}.
The history-agnostic question encoding $q_t$ is then fused with the history 
vector $\hat{h}_t$ via a fully connected layer to give a history-aware
question encoding $\hat{q}_t$ \eqref{eq:prog-dec-1}, which is fed into the \textit{program decoder}.

\small
\begin{multicols}{2}
\begin{center}\textbf{Question Encoder}\end{center}
\vspace*{-10pt}
\begin{align}
	\{\hat{w}_{ti}\} &= \text{LSTM}(\{w_{ti}\})
						\label{eq:ques-enc-1}\\
	q_t &= \hat{w}_{tT} \nonumber
\end{align}

\begin{center}\textbf{History Memory}\end{center}
\vspace*{-10pt}
\begin{align}
	\hat{h}_i &= \text{LSTM}(h_i)
    					\label{eq:hist-mem-1}\\
    \beta_{ti} &= \text{Softmax}(q_t^T \hat{h}_i)
    					\nonumber\\
    \hat{h}_t &= \sum_{i=0}^{t-1} \beta_{ti} \hat{h}_i
    					\label{eq:hist-mem-2}\\
	\hat{q}_t &= \text{MLP}([q_t, \hat{h}_t])
    						\label{eq:prog-dec-1}           
\end{align} 
\vspace*{-20pt}
\columnbreak
\vspace*{-20pt}
\begin{center}\textbf{Program Decoder}\end{center}
\vspace*{-30pt}
\begin{align}
    \tilde u_{ti}^{(j)} &= \text{Linear}([\hat{w}_{tj}, d_{ti}])
    						\nonumber\\
    u_{ti}^{(j)} &= v^T \tanh(\tilde u_{ti}^{(j)})
    						\nonumber\\
    \alpha_{ti}^{(j)} &= \text{Softmax}(u_{ti}^{(j)})
    						\nonumber\\
    e_{ti} &= \sum_{j=1}^T \alpha_{ti}^{(j)} \hat{w}_{tj} 
							\label{eq:prog-dec-5}\\
    \tilde e_{ti} &= \text{MLP}([e_{ti}, d_{ti}]) 
							\label{eq:prog-dec-6}\\
    p(m_i | &\{m_k\}_{k=1}^{i-1}, Q_t, H) 
    						\nonumber\\
    	&= \text{Softmax}(\tilde e_{ti}) 
        					\label{eq:prog-dec-7}
\end{align} 
\end{multicols}
\normalsize

The decoder is another multi-layer LSTM network (with hidden states 
$\{d_{ti}\}$) which, at every time step $i$,
produces a soft attention map $\alpha_{ti}$ over the input sequence ($Q_t$) 
\cite{bahdanau_iclr15}.
This soft attention map for each module is used to compute the corresponding
text embedding, $x_{txt} = \sum_{j} \alpha_{ti}^{(j)} w_{tj}$.
Finally, to predict a module token $m_i$ at time step $i$, a weighted sum
of encoder hidden states $e_{ti}$ \eqref{eq:prog-dec-5} and the history-aware 
question vector $\hat{q}_t$ are combined via another fully-connected
layer \eqref{eq:prog-dec-6}, followed by a softmax to give a distribution 
$P(m_i | \{m_k\}_{k=1}^{i-1}, Q_t, H)$ over the module tokens 
\eqref{eq:prog-dec-7}.
During training, we minimize the cross-entropy loss 
$\mathcal{L}_Q^{prog}$ between this predicted distribution and the ground truth 
program tokens.
\Fig\ref{fig:model_fig} outlines the schematics of our program generator.

\myparagraph{Modules on captions.}
As the image caption $C$ is also a part of the dialog (history $H_0$ at round $0$),
it is desirable to track entities from $C$ via the coreference pool 
$P_{ref}$.
To this effect, we propose a novel extension of neural module networks to 
captions by using an auxiliary task that checks the alignment of a (caption, 
image) pair.
First, we learn to predict a program from $C$, different from those generated 
from $Q_t$, by minimizing the negative log-likelihood $\mathcal{L}_C^{prog}$,
akin to $\mathcal{L}_Q^{prog}$, of the ground truth caption program.
Next, we execute the caption program on two images $I^+ = I$ and $I^-$ 
(a random image from the dataset), to produce caption context vectors $c_C^+$
and $c_C^-$, respectively.
Note that $c_C^+$ and $c_C^-$ are different from the context vector $c_t$ 
produced from execution of the question program.
Finally, we learn a binary classifier on top to output classes $+1/-1$ for 
$c_C^+$ and $c_C^-$, respectively, by minimizing the binary cross entropy loss 
$\mathcal{L}_C^{aux}$.
The intuition behind the auxiliary task is:
to rightly classify aligned $(C, I^+)$ from misaligned $(C, I^-)$,
the modules will need to localize and focus on salient entities in the caption.
These entities (specifically, outputs from \texttt{Find} in the caption program)
are then collected in $P_{ref}$ for explicit coreference resolution on $Q_t$.

\myparagraph{Entities in answers.}
Using an analogous argument as above, answers from the previous rounds 
$\{A_i\}_{i=1}^{t-1}$ could have entities necessary to resolve coreferences in
$Q_t$.
For example,
\textit{`Q: What is the boy holding? A: A ball. Q: What color is it?'}
requires resolving `it' with the `ball' mentioned in the earlier answer.
To achieve this, at the end of round $t-1$, we encode 
$H_{t-1} = (Q_{t-1}, A_{t-1})$ as $h^{ref}_t$ using a multi-layer LSTM,
obtain the last image attention map $a$ fed to the last module in the program
that produced the context vector $c_t$, and add $(h^{ref}, a)$ as an additional 
candidate to the reference pool $P_{ref}$.
Notice that $h^{ref}$ contains the information about the answer $A_{t-1}$ in
the context of the question $Q_{t-1}$, while $a$ denotes the image attention
which was the last crucial step in arriving at $A_{t-1}$ in the earlier round.
In resolving coreferences in $Q_t$, if any, all the answers from previous
rounds now become potential candidates by virtue of being in $P_{ref}$.

\subsection{Other Model Components}
\label{sec:prog_execute}

\myparagraph{Program Execution.}
This component takes the generated program and associated text features 
$x_{txt}$ for each participating module, and executes it.
To do so, we first deserialize the given program from its RPN to a hierarchical
module layout.
Next, we arrange the modules dynamically according to the layout, giving us the 
network to answer $Q_t$.
At this point, the network is a simple feed-forward neural network, where we
start the computation from the leaf modules and feed outputs activations from 
modules at one layer as inputs into modules at the next layer 
(see \Fig\ref{fig:model_fig}).
Finally, we feed a context vector $c_t$ produced from the last module into
the next answer decoding component.

\myparagraph{Answer Decoding.}
\label{sec:ans_decode}
This is the last component of our model that uses the context vector $c_t$ to 
score answers from a pool of candidates $\mathcal{A}_t$, based on their
correctness.
The answer decoder: 
(a) encodes each candidate $A_t^{(i)}\in \mathcal{A}_t$ with a multi-layer LSTM
to obtain $o_t^{(i)}$,
(b) computes a score via a dot product with the context vector, \ie, $c_t^T o_t^{(i)}$, and
(c) applies a softmax activation to get a distribution over the candidates.
During training, we minimize the negative log-likelihood $\mathcal{L}_A^{dec}$
of the ground truth answer $A^{gt}_t$.
At test time, the candidate with the maximum score is picked as $\mathcal{A}_t$.
Using nomenclature from \cite{visdial}, this is a \textit{discriminative} 
decoder.
Note that our approach is not limited to a discriminative decoder, but can also 
be used with a \textit{generative} decoder (see supplement).
\myparagraph{Training Details.}
Our model components have fully differentiable operations within them.
Thus, to train our model, we combine the supervised loss terms from both
program generation 
$\{\mathcal{L}_Q^{prog}, \mathcal{L}_C^{prog}, \mathcal{L}_C^{aux}\}$
and answer decoding $\{\mathcal{L}^{dec}_A\}$, and minimize the sum total loss 
$\mathcal{L}^{total}$.

\section{Experiments}
\addresults{We  show results on the synthetic
MNIST Dialog dataset \cite{paul2017visual} with complex coreferences across rounds and a large %
dialog dataset on real 
images,  VisDial \cite{visdial}.}

We first show results on the synthetic
MNIST Dialog dataset \cite{paul2017visual}, 
designed to contain complex coreferences across rounds while being relatively 
easy textually and visually.
It is important to resolve these coreferences accurately in order to do 
well on this dataset, thus stress testing our model.
We then experiment with a large %
visual dialog dataset on real 
images,  VisDial \cite{visdial},
which offers both
linguistic and perceptual challenge in resolving visual coreferences and
grounding them in the image.
Implementation details are in the supplement.
\subsection{MNIST Dialog Dataset}

\noindent
\textbf{Dataset.}
The dialogs in the MNIST dialog dataset \cite{paul2017visual} are grounded in
images composed from a $4 \times 4$ grid of MNIST digits \cite{lecun-mnisthandwrittendigit-2010}.
Digits in the grid have four attributes---digit class ($0-9$), 
color, stroke, and background color.
Each dialog has $10$ question-answer pairs, where the questions are
generated through language templates, and the answers are single words.
Further, the questions are designed to query attributes of target digit(s),
count digits with similar attributes, \etc, all of which need tracking of the
target digits(s) by resolving references across dialog rounds.
Thus, coreference resolution plays a crucial part in the reasoning required to
answer the question, making the MNIST dataset both interesting and challenging
(\Fig\ref{fig:mnist_qualitative}).
The dataset contains $30k$ training, $10k$ validation, and $10k$ test images,
with three $10$-round dialogs for each image.

\begin{table}[t]
\begin{minipage}[b]{0.34\linewidth}
	\centering	
	\begin{tabular}[t]{lc}
	\toprule
	\textbf{Model} & \textbf{Acc.}\\
	\midrule
    I \cite{paul2017visual} & 20.2 \\
    Q \cite{paul2017visual} & 36.6 \\
    AMEM$\backslash$Seq \cite{paul2017visual} & 89.2 \\
    AMEM \cite{paul2017visual} & 96.4 \\
	NMN \cite{hu2017learning} & 23.8\\
    \cdashlinelr{1-2}
    \nmn{}$\backslash$Seq & 88.7 \\
    \nmn & \textbf{99.3} \\
	\bottomrule
\end{tabular}
    \vspace*{5pt}
    \caption{
    Answer accuracy on MNIST Dialog dataset.
    Higher the better.
    Our \nmn outperforms all other models with a near perfect
    accuracy on test set.}
    \label{tab:mnist_quantitative}
\end{minipage}\hfill
\begin{minipage}[t]{0.65\linewidth}
\centering
\includegraphics[width=\textwidth]{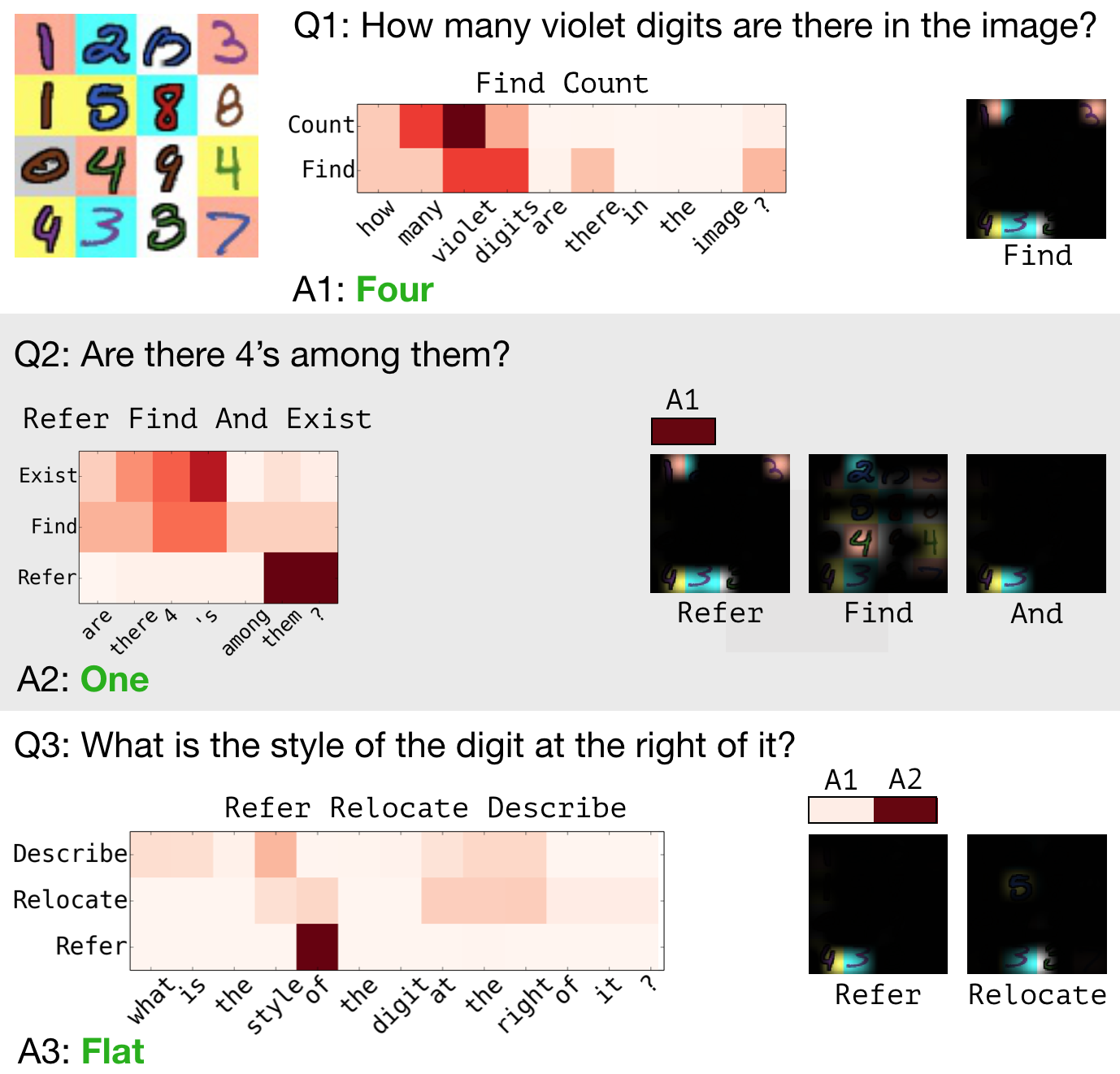}
\end{minipage}
\captionof{figure}{
  Illustration of explicit coreference resolution reasoning of our model on 
  the MNIST dialog dataset.
  For each question, a program and corresponding attentions ($\alpha$'s) 
  over question words (hot matrix on the left) is predicted.
  A layout is unpacked from the program, and modules are connected to
  form a feed-forward network used to answer the question, shown in green to
  indicate correctness.
  We also visualize output attention maps (right) from each participating 
  module.
  Specifically, in Q1 and Q2, \texttt{Find} localizes all violet digits and 
  4's, respectively (indicated by the corresponding $\alpha$).
  In Q2, \texttt{Refer} resolves \textit{`them'} and borrows the visual
  grounding from previous question.
}
\label{fig:mnist_qualitative}
\end{table}

\noindent
\textbf{Models and baselines.}
Taking advantage of single-word answers in this dataset, 
we simplify our answer decoder to be a $N$-way classifier, where $N$ is the number of possible answers.
Specifically, the context vector $c_t$ now passes through
a fully connected layer of size $N$, followed by softmax activations to give
us a distribution over possible answer classes.
At training time, we minimize the cross-entropy $\mathcal{L}_A^{dec}$ of the 
predicted answer distribution with the ground truth answer, at every round.
Note that single-word answers also simplify evaluation as answer accuracy can 
now be used to compare different models.
We further simplify our model by removing the memory augmentation to the
program generator, \ie, $\hat{q}_t = q_t$ \eqref{eq:prog-dec-1}, and denote it
as \nmn.
In addition to the full model, we also evaluate an ablation, 
\nmn{}$\backslash$Seq, without $\Delta_i t$ that additionally captured 
sequential nature of dialog (see \texttt{Refer} description).
We compete against the explicit reasoning model (NMN) \cite{hu2017learning} and
a comprehensive set of baselines AMEM, image-only (I), and question-only (Q), 
all from \cite{paul2017visual}.

\myparagraph{Supervision.}
In addition to the ground truth answer, we also need program supervision for 
questions to learn the program generation.
For each of the $5$ `types' of  questions, we manually create one program which
we apply as supervision for all questions of the corresponding type.
The type of question is provided with the question.
Note that our model needs program supervision only while training, and uses
predictions from program generator at test time.

\myparagraph{Results.}
\Table\ref{tab:mnist_quantitative} shows the results on MNIST dataset.
The following are the key observations:
(a) The text-only Q ($36.6\%$) and image-only I ($20.2\%$) do not perform well,
perhaps as expected as MNIST Dialog needs resolving strong coreferences to 
arrive at the correct answer.
For the same reason, NMN \cite{hu2017learning} has a low accuracy of $23.8\%$.
Interestingly, Q outperforms NMN by around $13\%$ (both use question and image,
but not history), possibly due to the explicit reasoning nature of NMN 
prohibiting it from capturing the statistic dataset priors.
(b) Our \nmn outperforms all other models with near perfect accuracy of $99.3\%$.
Examining the failure cases reveals that most of the mistakes made by \nmn was
due to misclassifying qualitatively hard examples from the original MNIST 
dataset.
(c) Factoring the sequential nature of the dialog additionally in the model is
beneficial, as indicated by the $10.6\%$ improvement in \nmn, and $7.2\%$ in
AMEM.
Intuitively, phrases with multiple potential referents, more often
than not, refer to the most recent referent, as seen in \Fig\ref{fig:teaser}, 
where \textit{`it'} has to be resolved to the closest referent in history.
\Fig\ref{fig:mnist_qualitative} shows a qualitative example.

\begin{table}[t]
	\centering
	\setlength{\tabcolsep}{6pt}
    \begin{tabular}{lcccccc}
	\toprule
	\textbf{Model} & \textbf{MRR} & \textbf{R$@$1} & \textbf{R$@$5}
    				& \textbf{R$@$10} & \textbf{Mean} \\%& \textbf{Grounding}\\
	\midrule
	MN-QIH-D \cite{visdial} 
    	& 0.597 & 45.55 & 76.22 & 85.37 & 5.46\\
    HCIAE-D-MLE \cite{lu_nips16} 
    	& 0.614 & 47.73 & 77.50 & 86.35 & 5.15\\
    AMEM+SEQ-QI \cite{paul2017visual} 
    	& 0.623 & 48.53 & 78.66 & 87.43 & 4.86\\
    NMN\cite{hu2017learning}
    	& 0.616 & 48.24 & 77.54 & 86.75 & 4.98\\
    \cdashlinelr{1-6}
	\nmn{}$\backslash$Mem
    	& 0.618 & 48.56 & 77.76 & 86.95 & 4.92 \\
	\nmn{}$\backslash\mathcal{L}_C^{aux}$
    	& \textbf{0.636} & \textbf{50.49} & 79.56 & 88.30 & 4.60\\
    \nmn{}$\backslash$Mem$\backslash\mathcal{L}_C^{aux}$
    	& 0.617 & 48.47 & 77.54 & 86.77 & 4.99 \\
    \nmn
    	& \textbf{0.636} & 50.24 & \textbf{79.81} & \textbf{88.51} & \textbf{4.53}\\
    \midrule
    \nmn (ResNet-152)
   	 	& 0.641 & 50.92 & 80.18 & 88.81 & 4.45\\
	\bottomrule
	\end{tabular}
    \caption{
    Retrieval performance on the validation set of VisDial v0.9 \cite{visdial} 
    (discriminative models) using VGG 	
    \cite{simonyan_iclr15} features (except last row).
    Higher the better for mean reciprocal rank (MRR) and recall$@k$ 
    (R$@1$, R$@5$, R$@10$), while lower the better for mean rank.
    Our \nmn model outperforms all other models across all metrics.}
    \label{tab:visdial_disc_result_table}
\end{table}

\subsection{VisDial v0.9 Dataset}

\noindent
\textbf{Dataset.}
The VisDial dataset \cite{visdial} is a crowd-sourced dialog dataset on COCO
images \cite{mscoco}, with free-form answers.
The publicly available VisDial v0.9 %
contains $10$-round dialogs on around $83k$ training images, and $40k$ 
validation images.
VisDial was collected from pairs of human workers, by instructing one of them 
to ask questions in a live chat interface to help them imagine the scene 
better.
Thus, the dialogs contain a lot of coreferences in natural language, which need 
to be resolved to answer the questions accurately.
\begin{figure}[t]
	\centering
    \includegraphics[width=0.9\textwidth]{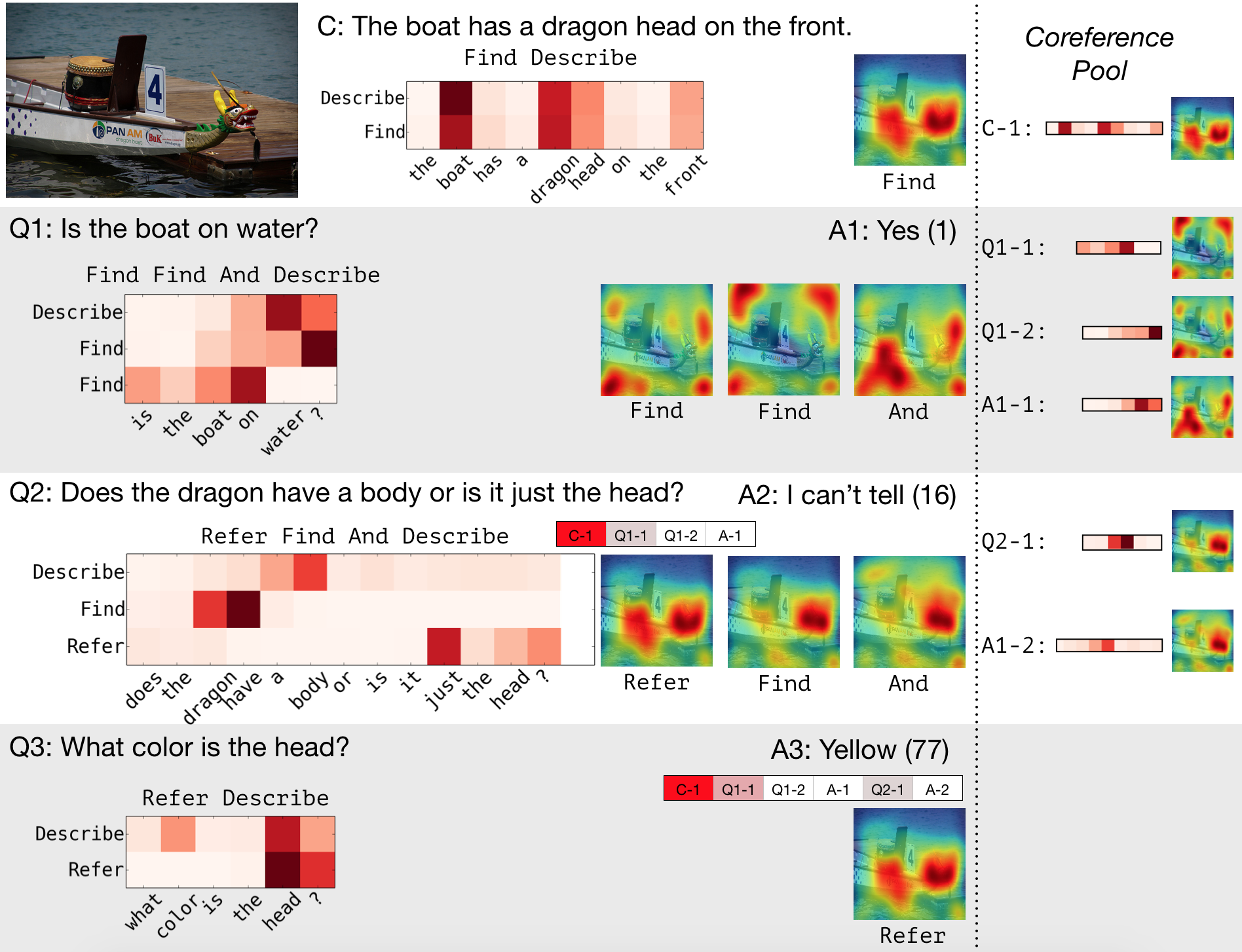}
    \caption{
    Example to demonstrate explicit coreference resolution by our \nmn model.
    It begins by grounding \textit{`dragon head'} from the caption C
    (shown on top), and saves it in the coreference pool $P_{ref}$ (right).
    At this point however, it does not consider the entity \textit{`boat'} 
    important, and misses it.
    Next, to answer Q1, it localizes \textit{`boat'} and \textit{`water'},
    both of which are `unseen', and rightly answers with \textit{Yes}.
    The ground truth rank (1 for Q1) is shown in the brackets.
    Additionally, it also registers these two entities in $P_{ref}$ for 
    coreference resolution in future dialog.
    For Q2, it refers the phrase \textit{`the head'} to the referent registered
    as C-1, indicated by attention on the bar above \texttt{Refer}.}
    \label{fig:visdial_qualitative}
\end{figure}

\myparagraph{Models and baselines.}
In addition to the \nmn model described in \Sec\ref{sec:approach}, 
we also consider ablations without the memory network augmented
program generator (\nmn{}$\backslash$Mem) or the auxiliary loss 
$\mathcal{L}_C^{aux}$ to train modules on captions 
(\nmn{}$\backslash\mathcal{L}_C^{aux}$), and without both
(\nmn{}$\backslash$Mem$\backslash\mathcal{L}_C^{aux}$).
As strong baselines, we consider: 
(a) neural module network without history \cite{hu2017learning} with answer
generation,
(b) the best \textit{discriminative} model based on memory networks MN-QIH-D 
from \cite{visdial}, %
(c) history-conditioned image attentive encoder (HCIAE-D-MLE) \cite{lu2017best},
and
(d) Attention-based visual coreference model (AMEM+SEQ-QI) \cite{paul2017visual}.
We use ImageNet pretrained VGG-16 \cite{simonyan_iclr15} to extract $x_{vis}$,
and also ResNet-152 \cite{he16cvpr} for \nmn{}.
Further comparisons are in supplement.

\myparagraph{Evaluation.}
Evaluation in visual dialog is via retrieval of the ground truth answer 
$A^{gt}_t$ from a pool of $100$ candidate answers 
$\mathcal{A}_t = \{A_t^{(1)}, \cdots A_t^{(100)}\}$.
These candidates are ranked based the discriminative decoder scores.
We report Recall$@k$ for $k=\{1, 5, 10\}$, mean rank, and mean reciprocal rank 
(MRR), as suggested by \cite{visdial},
on the set of $40k$ validation images (there is not test available for v0.9).

\myparagraph{Supervision.}
In addition to the ground truth answer $A^{gt}_t$ at each round, our model 
gets program supervision for $Q_t$, to train the program generator.
We automatically obtain (weak) program supervision from a language parser on 
questions (and captions) %
 \cite{hu16cvpr} %
and supervision to predict for \texttt{Refer} from %
an off-the-shelf text coreference resolution
tool\footnote{\url{https://github.com/huggingface/neuralcoref}}, based on
\cite{clark16emnlp}.
For questions that are a part of coreference chain, we replace \texttt{Find}
 with \texttt{Refer} in the parser supervised program. %
Our model predicts everything from the questions at test time. %

\myparagraph{Results.}
We summarize our observations from \Table\ref{tab:visdial_disc_result_table} below:
(a) Our \nmn outperforms all other approaches across all the metrics, 
	highlighting the importance of explicitly resolving coreferences for visual
	dialog.
  	Specifically, our R$@k$ ($k=1,2,5$) is at least 1 point higher than the 	
     best prior work (AMEM+SEQ-QI), and almost 2 points higher than NMN.
(b) Removing memory augmentation (\nmn{}$\backslash$Mem) hurts performance 
	uniformly over all metrics, as the model is unable to peek into 
    history to decide when to resolve coreferences via the \texttt{Refer} 
    module.
	Modules on captions seems to have varied effect on the full model, with
    decrease in R$@1$, but marginal increase or no effect in other metrics.
(c) \Fig\ref{fig:visdial_qualitative} illustrates the interpretable and 	
    grounded nature of our model.
\subsection{VisDial v1.0 Dataset}
\textbf{Dataset.}
Das \etal \cite{visdial} recently released VisDial v1.0 dataset.
Specifically, VisDial v1.0 comprises of:
(a) A re-organization of train and validation splits from v0.9 to form the new train v1.0.
Thus, train v1.0 now contains $120k$ images with $10-$round dialogs for each images, resulting
in a total of $1.2$ million question-answer pairs.
(b) An additional 10k COCO-like images from Flickr, on which crowd-sourced dialogs between
pairs of humans were collected similar to v0.9.
The 10k images are further split into 2k validation (val v1.0) and 8k test sets (test-std v1.0).
Dense candidate option annotations, which indicate the correctness 
of each candidate in the pool, were also collected for these 10k images.
Each image in the val v1.0 split is associated with a $10-$round dialog, 
while an image in test-std v1.0 has a variable-round dialog.

\noindent
\textbf{Additional Metrics and Models.}
Just as in the previous version (v0.9), the performance on the VisDial v1.0 dataset 
is benchmarked using standard retrieval metrics like Recall$@k$ ($k=\{1, 5, 10\}$),
mean reciprocal rank (MRR) and mean rank.
Further, Das \etal \cite{visdial} also propose to use normalized 
discounted cumulative gain (NDCG) to score the sorted pool of candidate
answers,
to evaluate on VisDial v1.0\footnote{\url{https://visualdialog.org/challenge/2018}}.
Intuitively, NDCG penalizes accurate answers that appear lower in the 
sorted pool based on a logarithmic weighting scheme, normalizing for 
the number of accurate answers across instances.
We train our \nmn{} model on train v1.0 and report numbers on test-std v1.0 split.
We compare against LF-QIH-D, HRE-QIH-D, and MN-QIH-D from \cite{visdial}, and
out-of-the-box neural module network (NMN) \cite{hu2017learning}.
The LF-QIH-D, HRE-QIH-D, and MN-QIH-D use VGG \cite{simonyan_iclr15} 
image features, while the neural module based models (\nmn{} and NMN) 
use ResNet-152 \cite{he16cvpr} features.

\noindent
\textbf{Results.}

Performance on VisDial v1.0 is given in \reftab{tab:visdial_v1_result_table}.
Our \nmn{} outperforms all other approaches on all metrics, except the neural module
baseline (NMN) on the NDCG metric.
We note that recent state-of-the-art, as reported on the leaderboard\footnote{https://evalai.cloudcv.org/web/challenges/challenge-page/103/leaderboard/298}, 
has reached up to 0.578 (NDCG) from team DL-61, but the approach and 
other details (\eg. features, use of an ensemble) are not fully known and unpublished at 
this point in time.

\begin{table}[t]
	\centering
	\setlength{\tabcolsep}{6pt}
    \begin{tabular}{lccccccc}
	\toprule
	\textbf{Model} & \textbf{MRR} & \textbf{R$@$1} & \textbf{R$@$5}
    				& \textbf{R$@$10} & \textbf{Mean} & \textbf{NDCG}\\
	\midrule
	LF-QIH-D \cite{visdial} (VGG) & 0.554 & 40.95 & 72.45 & 82.83 & 5.95 & 0.453 \\
    HRE-QIH-D \cite{visdial} (VGG) & 0.542 & 39.93 & 70.45 & 81.50 & 6.41 & 0.455\\
    MN-QIH-D \cite{visdial} (VGG) & 0.555 & 40.98 & 72.30 & 83.30 & 5.92 & 0.475 \\
	\cdashlinelr{1-7}
    NMN \cite{hu2017learning} & 0.588 & 44.15 & 76.88 & 86.88 & 4.81 & \textbf{0.581} \\
	\cdashlinelr{1-7}
    \nmn & \textbf{0.615} & \textbf{47.55} & \textbf{78.10} & \textbf{88.80} & \textbf{4.40} & 0.547 \\
    \bottomrule
	\end{tabular}
    \caption{
    Retrieval performance on the test-standard split of VisDial v1.0
    dataset \cite{visdial} (discriminative models).
    Higher the better for mean reciprocal rank (MRR), recall$@k$ 
    (R$@1$, R$@5$, R$@10$), and normalized discounted cumulative gain (NDGC) 
    while lower the better for mean rank.
    Our \nmn model outperforms all other models across all metrics, except neural
    module baseline (NMN) on NDGC.}
    \label{tab:visdial_v1_result_table}
\end{table}

\section{Conclusions}
We introduced a novel model for visual dialog based on neural module networks
that provides an introspective reasoning about visual coreferences.
It explicitly links coreferences and grounds them in the 
image at a word-level, rather than implicitly or at a sentence-level, as in prior visual dialog work.
Our \nmn outperforms prior work on both the MNIST dialog dataset (close to perfect accuracy), and on VisDial dataset, while being more interpretable, grounded, and consistent by construction.

\begingroup
\footnotesize
\myparagraph{Acknowledgements.}
This work was supported in part by NSF, AFRL, DARPA, Siemens, Google, Amazon, 
ONR YIPs and ONR Grants N00014-16-1-\{2713,2793\}, N000141210903. 
The views and conclusions contained herein are those of the authors and should 
not be interpreted as necessarily representing the official policies or 
endorsements, either expressed or implied, of the U.S. Government, or any 
sponsor.
\endgroup
\newpage
\appendix
\section*{Overview of Supplement}
\begin{figure}[t]
	\centering
    \includegraphics[width=\textwidth]{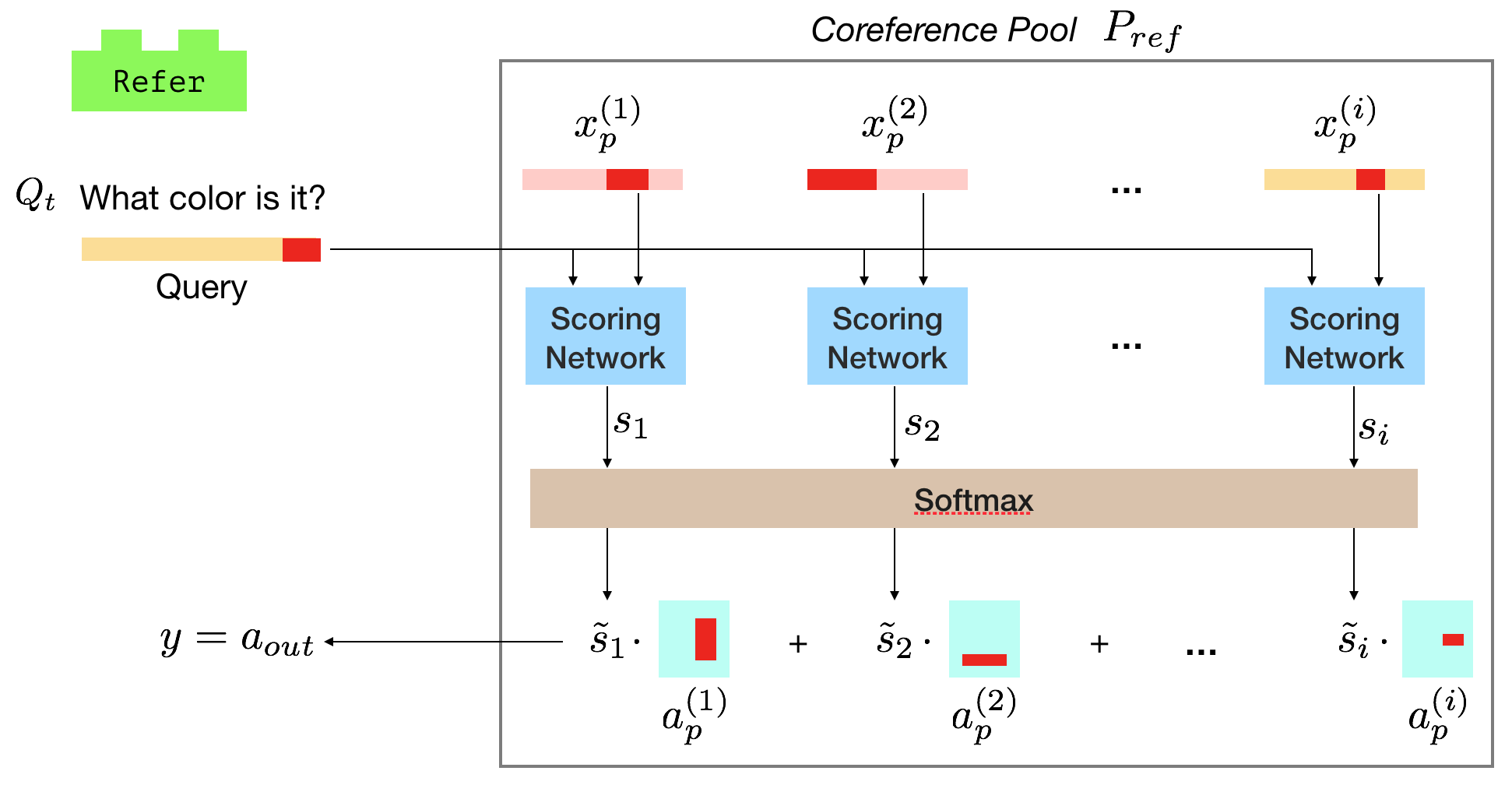}
	\caption{Visualization how our refer module accesses the coreference pool, to understand ``it'' in this example. For notation see main paper Section 3. $x_p$ can be seen as the keys to retrieve attentions $a_p$ from the coreference pool $P_{ref}$. The yellow and and pink lines symbolize the sentences/questions, and red is the text attention. Cyan boxes symbolize the image, with red being the spatial attention.}
    \label{fig:refer_visual}
\end{figure}

\begin{figure}
	\centering
    \includegraphics[width=0.9\textwidth]{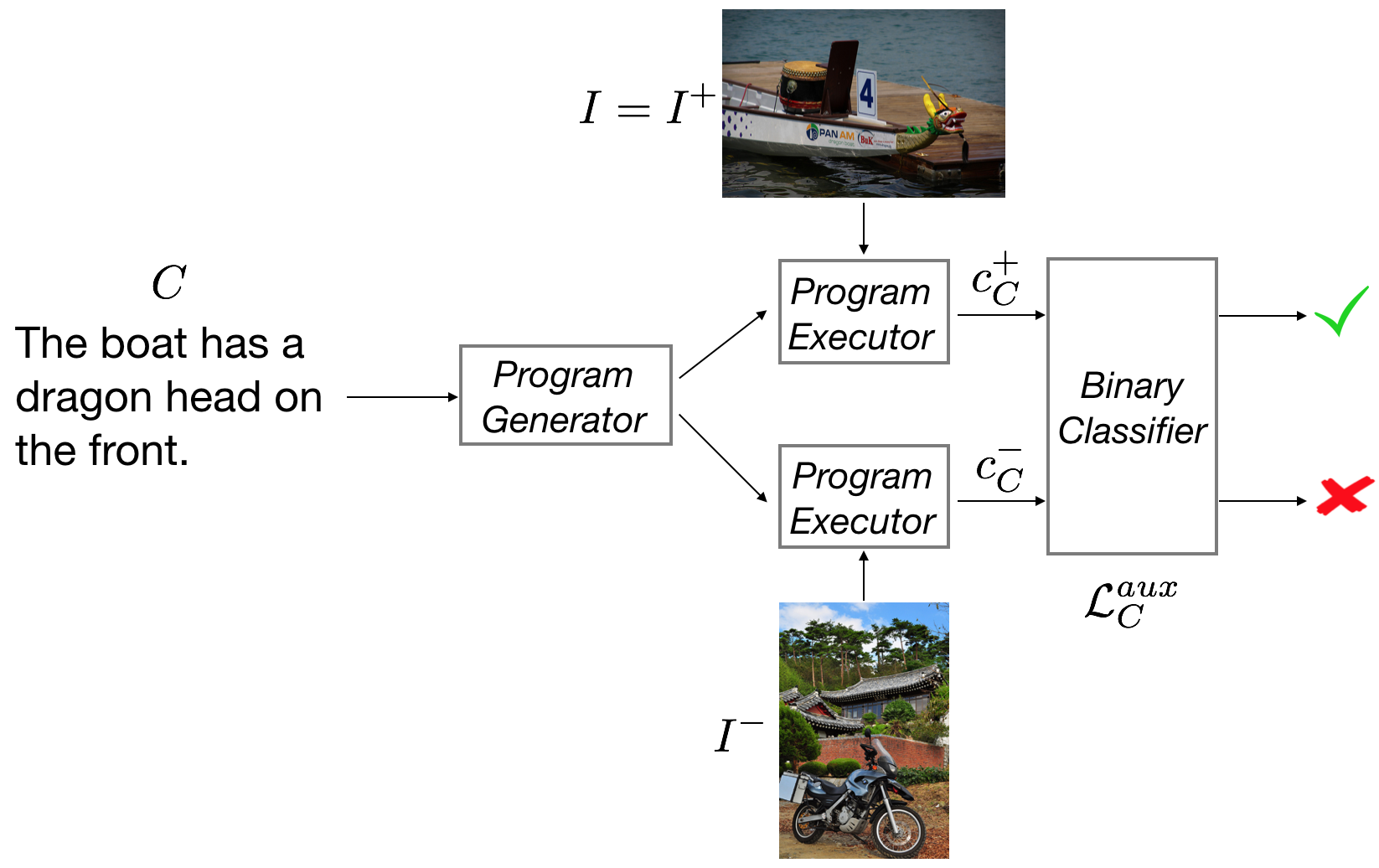}
    \caption{
    Flow diagram for the auxiliary task that measures the alignment between a
    caption and image as a binary classification task.
    By predicting a program for the caption, we propose a novel way to
    process captions at a finer word-level.
    Sec. 3.3 of the main paper motivates the use of modules on 
    captions via the auxiliary task.}
    \label{fig:nmn_caption_supp}
\end{figure}

\begin{figure}
	\centering
    \includegraphics[width=0.98\textwidth]{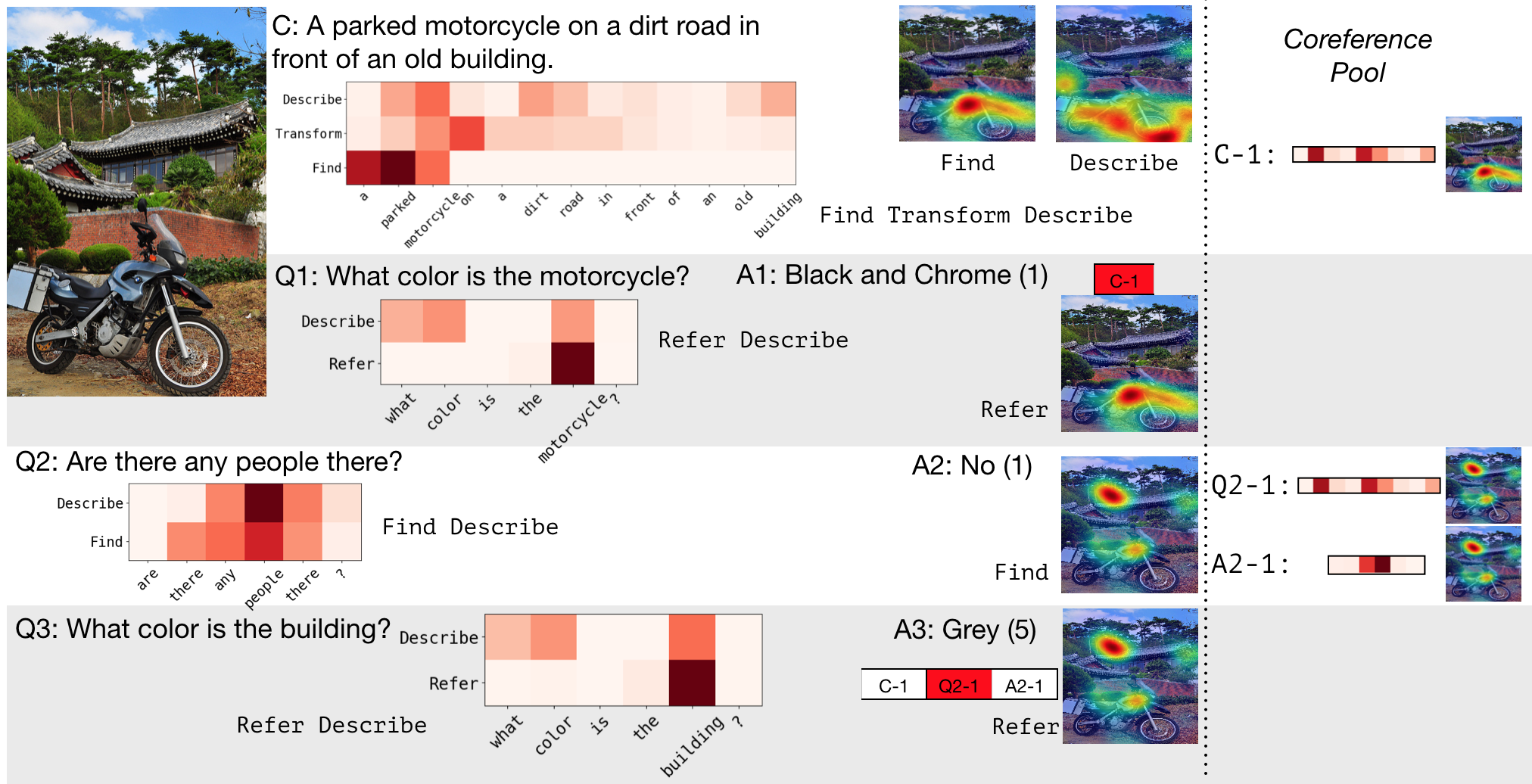}
    \caption{
    Qualitative example on VisDial dataset.}
    \label{fig:visdial_qual_2}
\end{figure}

The supplement is organized as follows:
\begin{itemize}
\setlength\itemsep{1em}
\item \Sec\ref{sup:disc_results} shows the results of our model using a 
	discriminative decoder with image features extracted using ImageNet 
    pretrained ResNet-152 \cite{he16cvpr},
    showing superior performance of our 
	explicit coreference model \nmn,
\item \Sec\ref{sup:gen_results} details our experiments with a generative 
	answer decoder,
\item Implementation details for our experiments are given in 
	\Sec\ref{sup:implement_details}, and
\item Schematics of our novel \texttt{Refer} module are in 
	\Fig\ref{fig:refer_visual}, \Fig\ref{fig:nmn_caption_supp} visualizes 
    the auxiliary task used to run modules on captions, a novel way to handle 
    captions at a fine word-level granularity, and \Fig\ref{fig:visdial_qual_2}
    shows another qualitative example from VisDial.
\end{itemize}

\section{Discriminative Decoder Experiments}
\label{sup:disc_results}
\myparagraph{Comparisons with ResNet-152 features.}
As mentioned in Sec. 4.2 of the main paper, the models trained on
VisDial v0.9 used an ImageNet pretrained VGG \cite{simonyan_iclr15} to extract
the image features $x_{vis}$.
In this section, we present results where a pretrained ResNet-152 \cite{he16cvpr}
was used to obtain the image features for our \nmn model, in \reftab{tab:visdial_disc_result_table_resnet}.
For a fair comparison, we obtained performance metrics from the authors for few
of these baselines (MNQIH-G, LF-QH-G), and retrain NMN with ResNet-152 features.

\section{Generative Decoder Experiments}
\label{sup:gen_results}

The main paper describes a discriminative answer decoder (Sec.3.4) and presents results for the same (Sec. 4.2) on VisDial v0.9 dataset.
As a reminder, a discriminative decoder takes the context vector $c_t$ as an input, and scores candidate answers according to their correctness.
In other words, a discriminative decoder needs to be presented with a list of
candidate answers and cannot `generate' novel answers.
We now introduce a generative answer decoder and present results on VisDial 
v0.9 dataset.

\myparagraph{Generative Answer Decoder} is a language model composed of a 
multi-layer LSTM with $c_t$ as its initial state. During training, we minimize
the negative log-likelihood $\mathcal{L}^{dec}_A$ of the ground truth answer
$A^{gt}_t$ with respect to the model.
At test time, we use the decoder to score all candidates in the answer pool 
$\mathcal{A}_t$ by model log-likelihood, and rank them accordingly.
Note that this ranking of candidate answers is done to comply with the 
evaluation protocol of VisDial v0.9, and is not a limitation of generative
answer decoder, unlike the discriminative one.
Thus, the generative answer decoder can potentially be used to generate novel
answers to a given question in the visual dialog via language generation.

\myparagraph{Models and baselines.}
We denote our model, as described in \Sec\ref{sec:approach}, with
\nmn to indicate that the model uses history $H$.
We also consider ablations which do not have the memory network augmented
program generator (\nmn{}$\backslash$Mem), or, the auxiliary loss 
$\mathcal{L}_C^{aux}$ to train modules on captions 
(\nmn{}$\backslash\mathcal{L}_C^{aux}$), and a combination of both
(\nmn{}$\backslash$Mem$\backslash\mathcal{L}_C^{aux}$).
As strong baselines,
we consider: 
(a) neural module network without history \cite{hu2017learning} with answer generation,
(b) the best \textit{generative} model based on memory networks MN-QIH-G from 
\cite{visdial}, in addition to their LF-QIG-G and HRE-QIH-G models, and
(c) history-conditioned image attentive encoder (HCIAE-G-MLE) \cite{lu2017best}.
We do not consider the HCIAE-G-DIS model with perceptual loss \cite{lu2017best}
as its contribution is complementary to our model.
Our model uses  ImageNet pretrained ResNet-152 
 \cite{he16cvpr} to extract features for images $x_{vis}$, while
some of these baseline models originally use VGG-16 \cite{simonyan_iclr15}
features.
For a fair comparison, we obtained performance metrics from the authors for few
of these baselines (MN-QIH-G, LF-QH-G), and report the others (HCIAE-G-MLE) as 
is.

\begin{table}[t]
	\centering
	\setlength{\tabcolsep}{6pt}
    \begin{tabular}{lcccccc}
	\toprule
	\textbf{Model} & \textbf{MRR} & \textbf{R$@$1} & \textbf{R$@$5}
    				& \textbf{R$@$10} & \textbf{Mean} \\%& \textbf{Grounding}\\
	\midrule
    LF-QIH-D* \cite{visdial} 
    	& 0.591 & 44.91 & 75.68 & 84.92 & 5.55\\
    HRE-QIH-D* \cite{visdial}
    	& 0.586 & 44.86 & 74.35 & 83.86 & 5.81\\
    MN-QIH-D* \cite{visdial}
    	& 0.601 & 46.04 & 76.78 & 85.93 & 5.29\\
    NMN\cite{hu2017learning}
    	& 0.620 & 48.89 & 77.83 & 86.99 & 4.94\\
	\cdashlinelr{1-6}
    \nmn
    	& \textbf{0.641} & \textbf{50.92} & \textbf{80.18} & \textbf{88.81} & \textbf{4.45}\\    
    \bottomrule
	\end{tabular}
    \caption{
    Retrieval performance on the validation set of VisDial
    dataset v0.9 \cite{visdial} 
    (discriminative models with ResNet-152 \cite{he16cvpr} features).
    Higher the better for mean reciprocal rank (MRR) and recall$@k$ 
    (R$@1$, R$@5$, R$@10$), while lower the better for mean rank.
    Our \nmn model outperforms all other models across all metrics.
    *indicates numbers obtained from authors for models retrained on ResNet-152
    features.}
    \label{tab:visdial_disc_result_table_resnet}
\end{table}

\myparagraph{Results.}
We summarize our observations from \Table\ref{tab:visdial_result_table} below:
(a) Our \nmn outperforms all other approaches according to R$@5$, R$@10$,
  and mean rank metrics, highlighting the importance of explicitly resolving
  coreferences for visual dialog.
  Specifically, our mean rank of $15.69$ is a $4\%$ improvement over the NMN 
  baseline.
(b) However, the NMN baseline has a higher R$@1$, and perhaps as a result,
	the best MRR.
    A possible reason could be due to the noisy supervision from the 
    out-of-domain, automatic, text-based coreference tool, as we entirely rely
    on it to predict \texttt{Refer}, the module responsible for coreference
    resolution.
(c) Our novel way of handling captions using modules (indicated
	$\mathcal{L}_C^{aux}$) boosts the full model, while the 
    hurting the ablation without the memory augmentation.
    That is, \nmn{}$\backslash$Mem$\backslash\mathcal{L}_C^{aux}$ is better 
    than \nmn{}$\backslash$Mem across all metrics, while 
    \nmn{}$\backslash \mathcal{L}_C^{aux}$ is uniformly worse than \nmn.

\begin{table}[t]
	\centering
	\setlength{\tabcolsep}{6pt}
    \begin{tabular}{lcccccc}
	\toprule
	\textbf{Model} & \textbf{MRR} & \textbf{R$@$1} & \textbf{R$@$5}
    				& \textbf{R$@$10} & \textbf{Mean} \\%& \textbf{Grounding}\\
	\midrule
	MN-QIH-G \cite{visdial} (VGG) & 0.526 & 42.29 & 62.85 & 68.88 & 17.06 \\
    HCIAE-G-MLE\cite{lu_nips16}(VGG) & 0.539 & 44.06 & \textbf{63.55} & 69.24 & 16.01\\
	\cdashlinelr{1-6}
    LF-QIH-G* \cite{visdial} & 0.515 & 41.04 & 61.63 & 67.54 & 17.32\\
    HRE-QIH-G* \cite{visdial} & 0.523 & 42.26 & 62.20 & 67.95 & 16.96\\
    MN-QIH-G* \cite{visdial} & 0.527 & 42.60 & 62.58 & 68.52 & 17.21\\
    NMN\cite{hu2017learning} & \textbf{0.542} & \textbf{45.05} & 63.27 & 69.28 & 16.34\\
	\cdashlinelr{1-6}
    \nmn{}$\backslash$Mem & 0.531 & 43.67 & 62.31 & 68.27 & 16.73 \\
    \nmn{}$\backslash$Mem$\backslash\mathcal{L}_C^{aux}$ 
				& 0.537 & 44.26 & 63.02 & 69.01 & 16.47 \\
	\nmn{}$\backslash\mathcal{L}_C^{aux}$ &
    			0.533 & 43.62 & 63.08 & 69.12 & 16.39 \\
    \nmn & 0.535 & 43.66 & \textbf{63.54} & \textbf{69.93} & \textbf{15.69}\\
    \bottomrule
	\end{tabular}
    \caption{
    Retrieval performance on the validation set of VisDial v0.9 
    \cite{visdial} (generative models).
    Higher the better for mean reciprocal rank (MRR) and recall$@k$ 
    (R$@1$, R$@5$, R$@10$), while lower the better for mean rank.
    Our \nmn model outperforms all other models in R$@5$, R$@10$, and mean rank.
    However, the neural module network baseline (NMN) has the best R$@1$, and 
    perhaps as a result, the best MRR as well.
    *indicates numbers obtained from authors for models retrained on 
    ResNet-152 features. \addresults{The right most column shows grounding performance (higher is better)}}
    \label{tab:visdial_result_table}
\end{table}


\section{Implementation Details}
\label{sup:implement_details}

All our models are implemented with Tensorflow v1.0 
\cite{tensorflow2015-whitepaper}.
To optimize, we use Adam \cite{kingma2014adam} with a learning rate of $0.0001$.
Gradients at each iteration are clamped to $[-2.0, 2.0]$ to avoid gradient explosion.
To preprocess text, we follow \cite{visdial}, \ie, we lowercase all questions 
and answers, and tokenize using the Python NLTK framework \cite{nltk}.
We then construct a dictionary of all words that appear at least five times in
the training set.
Specific model hyperparameters for each experiment are given below.

\subsection{MNIST Dialog Dataset}

Due to the synthetic nature, the text in the dialog is of low variability 
and is made up of a small vocabulary.
Thus, we only use a single-layered LSTM with a hidden size of $64$ to encode 
both questions and history.
For each word in our vocabulary of size $73$, we learn embeddings with $32$
dimensions.
We learn a similar dimensional embedding for each of our modules.
To extract image features, we design a convolutional neural network (CNN) with
the same architecture as \cite{paul2017visual}.
Specifically, our CNN has four $3 \times 3$ convolutional layers, 
each followed by a batch norm, ReLU non-linearity, and a $2 \times 2$ max pool
layer.
While the first two convolutional layers have $32$ feature channels, the last 
two have $64$ channels each.
To pick the best model, we use early stoppage on the provided validation
set of $10k$ images.

\subsection{VisDial Dataset}

Each LSTM used in our VisDial experiments has two layers and with a hidden size
of $1000$.
To represent images, we use convolutional features before the final mean pooling
from a ImageNet pre-trained ResNet-152 \cite{he16cvpr} model.
Further, we also add two additional dimensions indicating the $X$ (columns) and 
$Y$ (rows) locations respectively, to facilitate the model in handling
spatial reasoning.
With a large vocabulary of around $8k$ words, our word and module embeddings 
are $300$ dimensional vectors.
We also initialize our word embeddings with GloVe \cite{pennington2014glove}.
We pick the best model via early stopping using mean
reciprocal rank metric on a subset of $3k$ images, set aside from the
$83k$ training images of VisDial v0.9.

\subsection{Document Changelog}
To help the readers track changes to this document, a brief changelog 
describing the revisions is provided below:

\begin{description}
\item{\textbf{v0:}} ECCV 2018 camera-ready version (not on arXiv).
\item{\textbf{v1:}} Initial arXiv version: Added experiments on VisDial v1.0 dataset.
\end{description}
\vspace*{-10pt}

\small
\bibliographystyle{splncs04}
\bibliography{biblioLong,rohrbach,main}

\begin{thebibliography}{10}
\providecommand{\url}[1]{\texttt{#1}}
\providecommand{\urlprefix}{URL }
\providecommand{\doi}[1]{https://doi.org/#1}

\bibitem{nltk}
{NLTK}. \url{http://www.nltk.org/}

\bibitem{tensorflow2015-whitepaper}
Abadi, M., Agarwal, A., Barham, P., Brevdo, E., Chen, Z., Citro, C., Corrado,
  G.S., Davis, A., Dean, J., Devin, M., Ghemawat, S., Goodfellow, I., Harp, A.,
  Irving, G., Isard, M., Jia, Y., Jozefowicz, R., Kaiser, L., Kudlur, M.,
  Levenberg, J., Man\'{e}, D., Monga, R., Moore, S., Murray, D., Olah, C.,
  Schuster, M., Shlens, J., Steiner, B., Sutskever, I., Talwar, K., Tucker, P.,
  Vanhoucke, V., Vasudevan, V., Vi\'{e}gas, F., Vinyals, O., Warden, P.,
  Wattenberg, M., Wicke, M., Yu, Y., Zheng, X.: {TensorFlow}: Large-scale
  machine learning on heterogeneous systems (2015),
  \url{https://www.tensorflow.org/}, software available from tensorflow.org

\bibitem{AgrawalBP16}
Agrawal, A., Batra, D., Parikh, D.: Analyzing the behavior of visual question
  answering models. In: Proceedings of the Conference on Empirical Methods in
  Natural Language Processing (EMNLP) (2016)

\bibitem{anderson18cvpr}
Anderson, P., He, X., Buehler, C., Teney, D., Johnson, M., Gould, S., Zhang,
  L.: Bottom-up and top-down attention for image captioning and vqa. In:
  Proceedings of the IEEE Conference on Computer Vision and Pattern Recognition
  (CVPR) (2018)

\bibitem{andreas_naacl16}
Andreas, J., Rohrbach, M., Darrell, T., Klein, D.: {Learning to Compose Neural
  Networks for Question Answering} (2016)

\bibitem{andreas16cvpr}
Andreas, J., Rohrbach, M., Darrell, T., Klein, D.: Neural module networks. In:
  Proceedings of the IEEE Conference on Computer Vision and Pattern Recognition
  (CVPR) (2016)

\bibitem{hendricks17iccv}
Anne~Hendricks, L., Wang, O., Shechtman, E., Sivic, J., Darrell, T., Russell,
  B.: Localizing moments in video with natural language. In: Proceedings of the
  IEEE International Conference on Computer Vision (ICCV) (2017)

\bibitem{antol15iccv}
Antol, S., Agrawal, A., Lu, J., Mitchell, M., Batra, D., Zitnick, C.L., Parikh,
  D.: Vqa: Visual question answering. In: Proceedings of the IEEE International
  Conference on Computer Vision (ICCV) (2015)

\bibitem{bahdanau_iclr15}
Bahdanau, D., Cho, K., Bengio, Y.: {Neural Machine Translation by Jointly
  Learning to Align and Translate}. In: Proceedings of the International
  Conference on Learning Representations (ICLR) (2015)

\bibitem{bergsma06acl}
Bergsma, S., Lin, D.: Bootstrapping path-based pronoun resolution. In:
  Proceedings of the Annual Meeting of the Association for Computational
  Linguistics (ACL). Association for Computational Linguistics (2006)

\bibitem{10.2307/2001990}
Burks, A.W., Warren, D.W., Wright, J.B.: An analysis of a logical machine using
  parenthesis-free notation. Mathematical Tables and Other Aids to Computation
  \textbf{8}(46),  53--57 (1954), \url{http://www.jstor.org/stable/2001990}

\bibitem{clark16emnlp}
Clark, K., Manning, C.D.: Deep reinforcement learning for mention-ranking
  coreference models. In: Proceedings of the Conference on Empirical Methods in
  Natural Language Processing (EMNLP) (2016)

\bibitem{visdial}
Das, A., Kottur, S., Gupta, K., Singh, A., Yadav, D., Moura, J.M., Parikh, D.,
  Batra, D.: {V}isual {D}ialog. In: CVPR (2017)

\bibitem{visdial_rl}
Das, A., Kottur, S., Moura, J.M., Lee, S., Batra, D.: {L}earning {C}ooperative
  {V}isual {D}ialog {A}gents with {D}eep {R}einforcement {L}earning. arXiv
  preprint arXiv:1703.06585  (2017)

\bibitem{fukui16emnlp}
Fukui, A., Park, D.H., Yang, D., Rohrbach, A., Darrell, T., Rohrbach, M.:
  Multimodal compact bilinear pooling for visual question answering and visual
  grounding. In: Proceedings of the Conference on Empirical Methods in Natural
  Language Processing (EMNLP) (2016)

\bibitem{geman_pnas14}
Geman, D., Geman, S., Hallonquist, N., Younes, L.: {Visual Turing Test for
  computer vision systems}. In: Proceedings of the National Academy of Sciences
  (2015)

\bibitem{goyal17cvpr}
Goyal, Y., Khot, T., Summers-Stay, D., Batra, D., Parikh, D.: Making the v in
  vqa matter: Elevating the role of image understanding in visual question
  answering. In: Proceedings of the IEEE Conference on Computer Vision and
  Pattern Recognition (CVPR) (2017)

\bibitem{grice1975logic}
Grice, H.P.: Logic and conversation. In: Cole, P., Morgan, J.L. (eds.) Syntax
  and Semantics: Vol. 3: Speech Acts, pp. 41--58. Academic Press, New York
  (1975), \url{http://www.ucl.ac.uk/ls/studypacks/Grice-Logic.pdf}

\bibitem{he16cvpr}
He, K., Zhang, X., Ren, S., Sun, J.: Deep residual learning for image
  recognition. In: Proceedings of the IEEE Conference on Computer Vision and
  Pattern Recognition (CVPR) (2016)

\bibitem{hu2017learning}
Hu, R., Andreas, J., Rohrbach, M., Darrell, T., Saenko, K.: Learning to reason:
  End-to-end module networks for visual question answering. In: Proceedings of
  the IEEE International Conference on Computer Vision (ICCV) (2017)

\bibitem{hu16cvpr}
Hu, R., Xu, H., Rohrbach, M., Feng, J., Saenko, K., Darrell, T.: Natural
  language object retrieval. In: Proceedings of the IEEE Conference on Computer
  Vision and Pattern Recognition (CVPR) (2016)

\bibitem{JohnsonHMFZG16}
Johnson, J., Hariharan, B., van~der Maaten, L., Fei-Fei, L., Zitnick, C.L.,
  Girshick, R.: Clevr: A diagnostic dataset for compositional language and
  elementary visual reasoning. In: Proceedings of the IEEE Conference on
  Computer Vision and Pattern Recognition (CVPR). IEEE (2017)

\bibitem{johnson17iccv}
Johnson, J., Hariharan, B., van~der Maaten, L., Hoffman, J., Fei-Fei, L.,
  Zitnick, C.L., Girshick, R.: Inferring and executing programs for visual
  reasoning. In: Proceedings of the IEEE International Conference on Computer
  Vision (ICCV) (2017)

\bibitem{kingma2014adam}
Kingma, D., Ba, J.: Adam: A method for stochastic optimization. arXiv:1412.6980
   (2014)

\bibitem{kong14cvpr}
Kong, C., Lin, D., Bansal, M., Urtasun, R., Fidler, S.: What are you talking
  about? text-to-image coreference. In: Proceedings of the IEEE Conference on
  Computer Vision and Pattern Recognition (CVPR) (2014)

\bibitem{lecun-mnisthandwrittendigit-2010}
LeCun, Y., Cortes, C.: {MNIST} handwritten digit database  (2010),
  \url{http://yann.lecun.com/exdb/mnist/}

\bibitem{lin14cvpr}
Lin, D., Fidler, S., Kong, C., Urtasun, R.: Visual semantic search: Retrieving
  videos via complex textual queries. In: Proceedings of the IEEE Conference on
  Computer Vision and Pattern Recognition (CVPR) (2014)

\bibitem{mscoco}
Lin, T.Y., Maire, M., Belongie, S., Hays, J., Perona, P., Ramanan, D., Dollár,
  P., Zitnick, C.L.: {Microsoft COCO: Common Objects in Context}. In:
  Proceedings of the European Conference on Computer Vision (ECCV) (2014)

\bibitem{lu2017best}
Lu, J., Kannan, A., Yang, J., Parikh, D., Batra, D.: Best of both worlds:
  Transferring knowledge from discriminative learning to a generative visual
  dialog model. In: Advances in Neural Information Processing Systems (NIPS)
  (2017)

\bibitem{lu_nips16}
Lu, J., Yang, J., Batra, D., Parikh, D.: {Hierarchical Question-Image
  Co-Attention for Visual Question Answering}. In: Advances in Neural
  Information Processing Systems (NIPS) (2016)

\bibitem{malinowski15iccv}
Malinowski, M., Rohrbach, M., Fritz, M.: Ask your neurons: A neural-based
  approach to answering questions about images. In: Proceedings of the IEEE
  International Conference on Computer Vision (ICCV) (2015)

\bibitem{mao16cvpr}
Mao, J., Huang, J., Toshev, A., Camburu, O., Yuille, A., Murphy, K.: Generation
  and comprehension of unambiguous object descriptions. In: Proceedings of the
  IEEE Conference on Computer Vision and Pattern Recognition (CVPR) (2016)

\bibitem{1802.03803}
Massiceti, D., Siddharth, N., Dokania, P.K., Torr, P.H.S.: Flipdial: A
  generative model for two-way visual dialogue (2018)

\bibitem{mitchell-vandeemter-reiter:2013:NAACL-HLT}
Mitchell, M., van Deemter, K., Reiter, E.: Generating expressions that refer to
  visible objects. In: Proceedings of the 2013 Conference of the North American
  Chapter of the Association for Computational Linguistics: Human Language
  Technologies. pp. 1174--1184. Association for Computational Linguistics,
  Atlanta, Georgia (June 2013), \url{http://www.aclweb.org/anthology/N13-1137}

\bibitem{pennington2014glove}
Pennington, J., Socher, R., Manning, C.D.: Glove: Global vectors for word
  representation. In: Proceedings of the Conference on Empirical Methods in
  Natural Language Processing (EMNLP) (2014)

\bibitem{plummer15iccv}
Plummer, B., Wang, L., Cervantes, C., Caicedo, J., Hockenmaier, J., Lazebnik,
  S.: Flickr30k entities: Collecting region-to-phrase correspondences for
  richer image-to-sentence models. In: Proceedings of the IEEE International
  Conference on Computer Vision (ICCV) (2015)

\bibitem{ramanathan14eccv}
Ramanathan, V., Joulin, A., Liang, P., Fei-Fei, L.: Linking people in videos
  with "their" names using coreference resolution. In: Proceedings of the
  European Conference on Computer Vision (ECCV) (2014)

\bibitem{regneri13tacl}
Regneri, M., Rohrbach, M., Wetzel, D., Thater, S., Schiele, B., Pinkal, M.:
  {Grounding Action Descriptions in Videos}. Transactions of the Association
  for Computational Linguistics (TACL)  \textbf{1},  25--36 (2013)

\bibitem{rohrbach16eccv}
Rohrbach, A., Rohrbach, M., Hu, R., Darrell, T., Schiele, B.: Grounding of
  textual phrases in images by reconstruction. In: Proceedings of the European
  Conference on Computer Vision (ECCV) (2016)

\bibitem{rohrbach17cvpr}
Rohrbach, A., Rohrbach, M., Tang, S., Oh, S.J., Schiele, B.: Generating
  descriptions with grounded and co-referenced people. In: Proceedings of the
  IEEE Conference on Computer Vision and Pattern Recognition (CVPR) (2017)

\bibitem{paul2017visual}
Seo, P.H., Lehrmann, A., Han, B., Sigal, L.: Visual reference resolution using
  attention memory for visual dialog. In: Advances in Neural Information
  Processing Systems (NIPS) (2017)

\bibitem{simonyan_iclr15}
Simonyan, K., Zisserman, A.: {Very deep convolutional networks for large-scale
  image recognition}. In: Proceedings of the International Conference on
  Learning Representations (ICLR) (2015)

\bibitem{DBLP:journals/corr/StrubVMPCP17}
Strub, F., de~Vries, H., Mary, J., Piot, B., Courville, A.C., Pietquin, O.:
  End-to-end optimization of goal-driven and visually grounded dialogue
  systems. In: Proceedings of the International Joint Conference on Artificial
  Intelligence (IJCAI) (2017)

\bibitem{guesswhat}
de~Vries, H., Strub, F., Chandar, S., Pietquin, O., Larochelle, H., Courville,
  A.C.: Guesswhat?! visual object discovery through multi-modal dialogue. In:
  Proceedings of the IEEE Conference on Computer Vision and Pattern Recognition
  (CVPR) (2017)

\bibitem{wang2016cvpr}
Wang, L., Li, Y., Lazebnik, S.: Learning deep structure-preserving image-text
  embeddings. In: Proceedings of the IEEE Conference on Computer Vision and
  Pattern Recognition (CVPR) (2016)

\bibitem{DBLP:journals/corr/WestonCB14}
Weston, J., Chopra, S., Bordes, A.: Memory networks. In: Proceedings of the
  International Conference on Learning Representations (ICLR) (2015)

\bibitem{winograd1971procedures}
Winograd, T.: Procedures as a representation for data in a computer program for
  understanding natural language. Tech. rep., DTIC Document (1971)

\bibitem{Winograd:1972:UNL:540414}
Winograd, T.: Understanding Natural Language. Academic Press, Inc., Orlando,
  FL, USA (1972)

\bibitem{yu13acl}
Yu, H., Siskind, J.M.: Grounded language learning from videos described with
  sentences. In: Proceedings of the Annual Meeting of the Association for
  Computational Linguistics (ACL) (2013)

\bibitem{yu16eccv}
Yu, L., Poirson, P., Yang, S., Berg, A.C., Berg, T.L.: Modeling context in
  referring expressions. In: Proceedings of the European Conference on Computer
  Vision (ECCV) (2016)

\bibitem{ZhangGSBP15}
Zhang, P., Goyal, Y., Summers-Stay, D., Batra, D., Parikh, D.: {Yin and Yang:
  Balancing and Answering Binary Visual Questions}. In: Proceedings of the IEEE
  Conference on Computer Vision and Pattern Recognition (CVPR) (2016)

\end{thebibliography}
\end{document}